\documentclass[preprint,review,12pt]{elsarticle}

\usepackage{amsmath,amsfonts,amssymb}
\usepackage{caption}
\usepackage{subcaption}
\usepackage{todonotes}
\usepackage{hyperref}

\journal{Applied Soft Computing}

\begin{document}

\begin{frontmatter}



\title{Continuous Cartesian Genetic Programming based representation  for 
Multi-Objective Neural Architecture Search}


\author{Cosijopii Garcia-Garcia}
\ead{cosijopii@inaoe.mx}
\author{Alicia Morales-Reyes}
\ead{a.morales@inaoep.mx}
\author{Hugo Jair Escalante}
\ead{hugojair@inaoep.mx}
\affiliation{organization={Computer science,  
INAOE},
            addressline={Luis Enrique Erro \#1}, 
            city={San Andres Cholula},
            postcode={72840}, 
            state={Puebla},
            country={Mexico}}
\begin{abstract}
We propose a novel approach for the challenge of designing less complex yet highly effective convolutional neural networks (CNNs) through the use of cartesian genetic programming (CGP) for neural architecture search (NAS).
Our approach combines real-based and block-chained CNNs representations based on CGP for optimization in the continuous domain using multi-objective evolutionary algorithms (MOEAs). Two variants are introduced that differ in the granularity of the search space they consider.  The proposed CGP-NASV1 and CGP-NASV2 algorithms were evaluated using the non-dominated sorting genetic algorithm II (NSGA-II) on the CIFAR-10 and CIFAR-100 datasets. The empirical analysis was extended to assess the crossover operator from differential evolution (DE), the multi-objective evolutionary algorithm based on decomposition (MOEA/D) and \textit{S metric selection} evolutionary multi-objective algorithm (SMS-EMOA)  using the same representation. Experimental results demonstrate that our approach is competitive with state-of-the-art proposals in terms of classification performance and model complexity.
 
\end{abstract}




\end{frontmatter}


\section{Introduction}
\label{intro}
 


Deep neural networks (DNNs), particularly convolutional  and recurrent neural networks, have recently gained considerable  popularity for approaching a wide variety of problems~\citep{Young2018,Kolbk2017,Grigorescu2019,Martinez2021}. Regarding convolutional neural networks (CNNs), these are very effective computational models that have been thoroughly investigated in a wide range of image processing and computer vision-related tasks. This has been possible thanks to a number of factors including availability of large amounts of data, high-performance computing resources and research advances in the machine learning field  \citep{Miikkulainen2019}.

CNNs are often organized according to complex topologies (architectures) whose design, including hyperparameter configuration, is done by expert users. 
While CNNs grow in complexity, manual configuration becomes time consuming and in some cases unfeasible~\citep{Miikkulainen2019}. Neural Architecture Search (NAS) is a field that aims at automating the design and configuration of CNN models~\citep{Elsken2019}. NAS methods explore the space of CNN topologies looking for an architecture that meets certain criteria, for instance, achieving a minimum performance, or being \emph{light} in terms of the number of parameters. 

Nowadays, progress in NAS research has resulted in the identification of novel CNNs with favorable performance on representative image classification datasets, attracting the scientific community's attention to this topic~\citep{Lu2020a}.

Evolutionary computation (EC) is a set of techniques inspired by the process of natural evolution to deal with complex optimization problems among others. The multi-objective approach deals with problems where objectives are in conflict and focuses on finding a set of solutions that represent a trade-off between them~\citep{Eiben2015}. EC techniques have been successfully applied in the multi-objective optimization problems domain. In the context of NAS, the automated design of high-performance and low-complexity network architectures remains an open issue. This relevant problem can be formulated as one of  multi-objective optimization~\citep{Lu,Lu2020a,Lu2020,Garcia-Garcia2022,Termritthikun2021}. 

The motivation for multi-objective NAS is 
the fact that in real-world scenarios more computationally efficient and accurate CNN models are required. Therefore, objectives such as latency, number of operations, or power consumption are optimized in addition to the standard error or accuracy objectives~\citep{Pinos123}. 
In this paper, we introduce two multi-objective NAS methods based on Cartesian Genetic Programming (CGP), which aim to design CNN architectures for image classification.
Our proposed CGP-NAS variants extend the original CGP-NAS~\cite{Garcia-Garcia2022} approach by considering a block-chain encoding. A first variant, called CGP-NASV1 is based on fixed CNN blocks to perform the NAS. While a second methodology, named CGP-NASV2, expands the search space to include the hyperparameters while relaxing the CGP-NASV1 fixed-blocks restriction. 
Our proposed NAS methods are evaluated in the context of image classification with CNNs. Experimental results in the CIFAR-10 and CIFAR-100 datasets show that both proposals are competitive against several state-of-the-art references. The evolved CNNs architectures have a lower number of parameters as well as lower complexity measured in Multiply-Adds (MAdds) operations in comparison to the state-of-the-art and maintain a low classification error.

A preliminary version of CGP-NAS was recently introduced in our previous work~\citep{Garcia-Garcia2022}. This NAS solution  
considers an encoding-decoding process, first using CGP to represent CNN architectures and then converting the CGP integer-based encoding to a real-based representation. 
A continuous representation allows to use canonical multi-objective evolutionary algorithms (MOEAs) for the optimization process. 
In fact, NSGA-II  was used as an optimizer for CGP-NAS in previous work~\citep{Garcia-Garcia2022}. 

This paper presents two novel evolutionary multi-objective NAS approaches based on our original CGP-NAS proposals. In this one, new features were added, such as the use of a high-level representation based on block-chain encoding, which allows for better control over the general shape of the architecture to be searched; also, new blocks of convolutional operations were added to the functions set; and finally, the representation was modified to have the feature of being able to evolve the hyperparameters together with the CNN architecture.

These models achieve competitive, state-of-the-art performance. Both enhancements are described in detail and extensively evaluated. NSGA-II\citep{Deb2002}, MOEA/D\citep{Zhang2007} and SMS-EMOA\citep{Nicala2007} are considered optimizers in our experimental evaluations.
 



The remainder of this paper is organized as follows. 
Section~\ref{sec:background} presents  background information required to understand our contribution. Next, Section 3 reviews related work focusing on NAS methods based on evolutionary computation. 
Section 4 introduces our proposed CGP-NAS variants. Then, Section 5 reports  the empirical assessment and Section 6 discusses the achieved results. Finally,  Section 7 outlines conclusions ad future work directions.

\section{Background}
\label{sec:background}
This section elaborates on the background information on Cartesian Genetic Programming (CGP) and solution representations for NAS.



\subsection{Cartesian Genetic Programming}


Cartesian Genetic Programming (CGP) is a Genetic Programming (GP) methodology based on acyclic graphs for solution representation that allow forward connections. This offers a number of improvements with respect to standard GP, where a tree-based representation is used. CGP was initially conceived to evolve digital circuits. It was proposed by J.F. Miller~\citep{MillerCGP}  and is called cartesian because it represents solutions using a two-dimensional grid~\citep{Miller2011a}. CGP shares important features with GP, such as the definition of a set of functions and their arity. CGP can also  be applied to different areas, such as the automatic design of digital circuits, mathematical equations, and even artistic applications~\citep{Miller2011a,MillerCGP}.

In the genotype space, a solution is mapped as an integer-based vector divided by segments that represent the function identifier (a predefined function from a set of functions that are the graph's nodes) and its connections. The size of each segment varies depending on the maximum arity of the represented function.
\begin{figure}[h]
    \centering
    \includegraphics[scale=0.95]{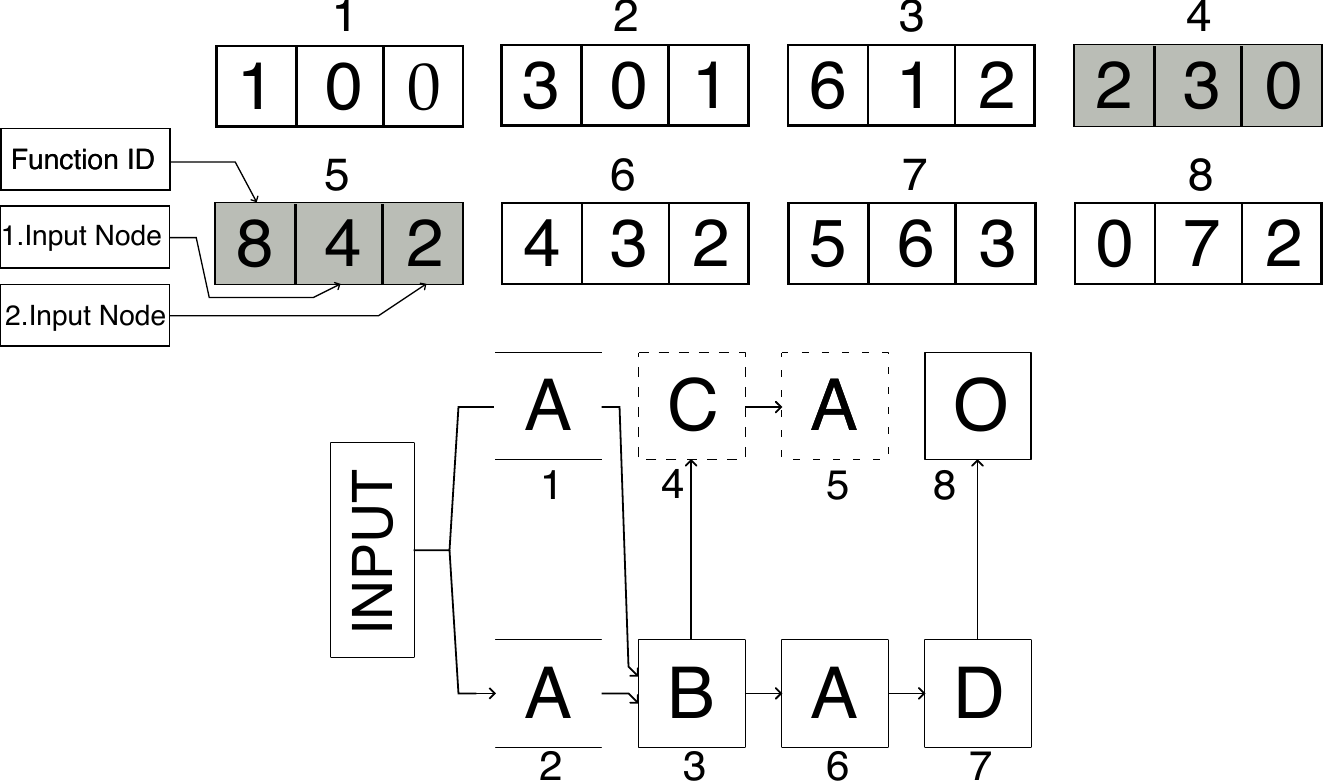}
    \caption{CGP representation. Top: integer-based genotype representation, Bottom: mapping to the phenotype space as an acyclic graph. Vectors in gray represent no connection, thus these are inactive nodes in the acyclic graph.}
    \label{fig:cgpexample}
\end{figure}

Figure \ref{fig:cgpexample} shows a CGP solution representation example with maximum arity of two and its mapping to the phenotype space.
Another CGP characteristic is the inactive nodes that are not expressed in the phenotype, these are represented as gray sections and dotted-line squares in Figure \ref{fig:cgpexample}. This information is maintained in the genotype and is exploited during the evolutionary process, for example, a change of an important connection between two functions maps as a big change in its corresponding phenotype. In canonic CGP, the crossover operator is not used (but can be implemented); instead, only mutations occur as random changes in connections and nodes.

The CGP is set up on a fixed-size grid $N_r \times N_c$, with the number of connections between nodes determined by an $l$ variable known as level-back. Normally, the number of inputs and outputs depends on the problem. CGP in its canonical version is combined with the $(1 + \lambda)$ evolutionary strategy algorithm. However, CGP can be adapted to other evolutionary search algorithms like NSGA-II or genetic algorithms~\citep{Miller2011a}.

\subsection{Solutions Representation in NAS}
\begin{figure}[h]
    \centering
    \includegraphics[scale=0.5]{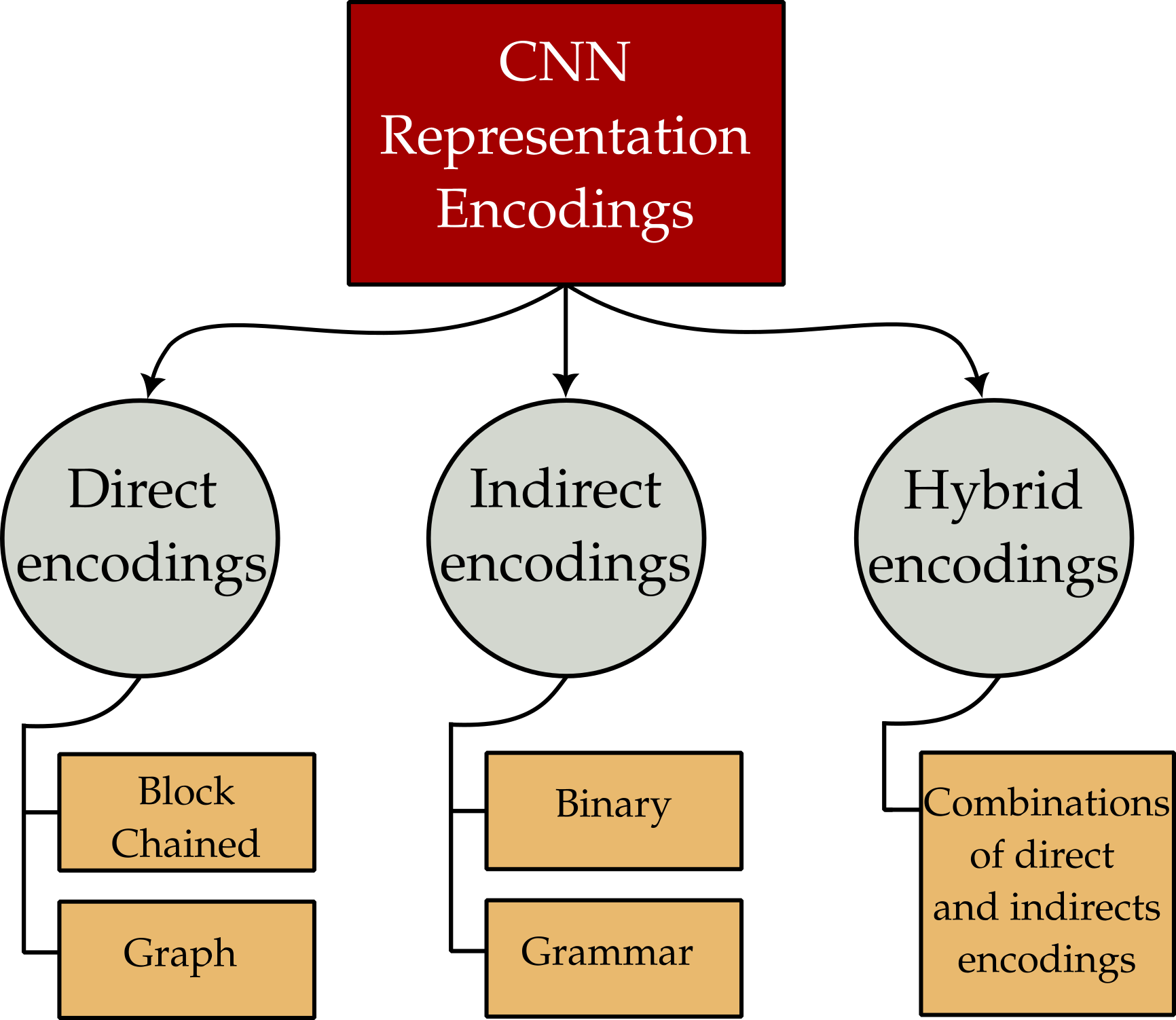}
    \caption{Representation types taxonomy to encode CNNs}
    \label{fig:taxo}
\end{figure}

NAS automates the design of neural networks given a specific task; for this, a search space of architectures must be specified, as well as a set of input data~\citep{Sun2023,Pinos123}. The complexity of the search space is indicated by the architecture representation. Neural architecture representation in NAS influences the design of the evolutionary algorithm as well as the complexity of the search space, and therefore it is of vital importance to understand the different ways in which the NAS representation can be classified. Figure~\ref{fig:taxo} shows a general taxonomy of encoding representations with three main categories~\citep{Vargas-Hakim2021}: direct, indirect, and hybrid encodings.


\textbf{Direct encoding} is based on graphs, trees, and block chains. Evolutionary operators struggle to adapt to this kind of encoding. A decoding mechanism is not necessary. 


\textbf{Indirect encoding}
relies on binary or integer representations; these are simpler and easy to adapt to standard evolutionary operators. However, they are less flexible and usually reduce the search space.

\textbf{Hybrid Encoding} is a combination of direct and indirect encoding. In the literature, a two-level block-chained representation is used on the first level and binary encoding on the second level. In the context of CGP-NAS~\citep{Garcia-Garcia2022}, CGP is used as a first-level encoding and real-based representation on a second level.

\section{Related Work}
The automated design of convolutional neural networks has been expanded by using neural architecture search. In a number of cases, the architectures discovered by NAS perform better than expert-tuned ones~\citep{Lu2020a}.
Commonly, NAS has been approached as a single-objective optimization problem, trying to minimize the classification error of architectures. However, several approaches have posed the NAS problem as a multi-objective optimization problem, targeting conflicting objectives such as the classification error, architecture complexity, and learnable parameters~\citep{Lu2020}. The following sections present an analysis of different single-objective and multi-objective NAS methodologies based on EC. Finishing with a discussion of different ways to represent the solutions.

\subsection{Single versus multi-objective NAS methods}
 NAS aims to automate the design process of neural network architectures. Several searching mechanisms have been used, such as evolutionary computation (EC), reinforcement learning (RL)~\citep{zoph2016neural,cui2019fast,cai2018proxylessnas}, as well as relaxation methods~\citep{liu2018darts,Chen_2019_ICCV,chuXiangxiang}.
Normally, NAS is divided into three main components: search space, search strategy, and the strategy to estimate the performance of each architecture. The first two components are linked to the representation and the search method; the strategy is most commonly used to estimate the performance of each architecture during the evolutionary search to obtain a training and validation subsets from the complete dataset, and therefore a performance estimate of the designed architecture~\citep{Elsken2019}.
Other performance estimation mechanisms, such as those based on 
Bayesian programming or surrogate models~\citep{Lu2020a,Lu,Lu2020}, have been used to reduce the total number of evaluations and thus the searching time. 

Most evolutionary single-objective NAS methods aim at decreasing the classification error rate, for example, Xie et al. introduced GeNet~\citep{Xie2017}, one of the first NAS models, where a straightforward model based on binary-fixed representation and a genetic algorithm was proposed. After GeNet, multiple proposals were introduced such as CGP-CNN~\citep{Suganuma2017} using Cartesian genetic programming (CGP) and a block-based design, AE-CNN~\citep{Sun2020} with the same idea of a block-based search but using a variable-length representation. 
DSGE-CNN~\citep{Lima2022} used the evolutionary grammar-based genetic programming algorithm with a block-based approach including the hyperparameters. This variant of Genetic Programming (GP), has become relevant because of the flexibility of the representation~\citep{Assuncao2019}. 

Tobari~\citep{Torabi2022} presented a new NAS algorithm that uses CGP, like CGP-CNN, which is a block-based algorithm that uses evolutionary strategies for the search. Its main contribution is to add a CGP crossover method based on the Multiple Sequence Alignment algorithm (MSA) inspired by the alignment of amino acid sequences, showing improvement in comparison to the canonical CGP.

Bakhshi~\citep{Bakhshi2020} proposed to add the hyperparameters to the search, all of these proposals have in common the use of predefined blocks.
CIFAR-10 and CIFAR-100 datasets were used for evaluation. 
The use of a variable-length representation similar to CGP-CNN, AE-CNN, DSGE-CNN, Tobari et al.'s  and Bakshi et al.'s solutions showed an improvement compared to initial proposals such as GeNet.



The automatic search for improved CNN architectures, in terms of better performance metrics or reduced complexity, inherently poses a multi-objective optimization problem. Including more architecture's elements within the search, such as complexity ( FLOPs or MAdds) or the number of learnable parameters~\citep{Termritthikun2021} prompted the area of Multi-objective Neural Architecture Search.
NEMO~\citep{Kim2017} was among the first proposals with a multi-objective approach. Later, NSGA-Net~\citep{Lu} stood out for its performance in finding CNN architectures with low error and complexity, followed by NSGA-NetV1~\citep{Lu2020a} and NSGA-NetV2~\citep{Lu2020}.

NSGA-Net uses binary representation in addition to a block-based approach with spatial reduction steps (pooling). On the other hand, NSGA-NetV1 uses the cell-based representation approach 
in which stacked blocks have internal connections between predefined blocks for convolution and pooling operations~\citep{zoph2018learning}. 
NSGA-Net extends the number of operations that can be performed, adding functions like dilated convolution and separable convolution. Thus, NSGA-netV1 uses graph-based encoding to represent solutions and both crossover and mutation operations. Finally, NSGA-NetV2 extends NSGA-NetV1 to a surrogate 
approach using integer-based encoding.


MOCNN encodes data using an Internet Protocol Address-based approach to define the number of parameters within a fixed DenseNet architecture and uses the Particle Swarm Optimization (PSO) algorithm for the architectures search~\citep{Wang2020}.
In MOGI-NET~\citep{Xue2021}, the encoding was inspired by NSGA-Net, the objectives are to minimize the classification error and the network parameters. It uses weight inheritance to reduce the number of epochs required to train the network, thus the offspring generated required less training.

EEEA-NET~\citep{Termritthikun2021} used a cell-based representation similar to NSGA-NetV1. It is based on a mechanism of early exit population initialization; a maximum threshold of learnable parameters is defined; if a randomly generated solution has a low number of parameters compared to the maximum threshold, it is added to the initial population; otherwise, another solution that meets this threshold is chosen. Authors intended, with this condition, to reduce the training time of each architecture, which impacts the total search time.

LF-MOGP~\citep{Liu2022} is a multi-objective proposal that uses a representation based on CGP and has as a major contribution the use of a leader-follower evolution mechanism. With this, during the evolutionary process, it keeps in an external archive the best solutions to generate diversity during the search. In its proposed evolutionary algorithm, it highlights the use of crossover and mutation, since in CGP, normally the crossover operators are not used. The objectives to be pursued in this work are the accuracy of the model as well as the complexity measured in the number of parameters.

EvoApproxNAS~\citep{Pinos123} is  a method based on CGP that targets three objectives in conflict: the power consumption of the processing platform, network parameters, and classification error. EvoApproxNAS increases the number of blocks for CGP by adding bottleneck residuals and inverted residual blocks.

Finally CGP-NAS~\citep{Garcia-Garcia2022} is a multi-objective approach based on a Cartesian genetic programming.
It minimizes the architecture complexity while reducing the classification error. CGP-NAS also combines CGP with a real-based encoding-decoding method; thus, solutions are manipulated in the continuous domain, and consequently, the classical multi-objective evolutionary algorithms can be used, in the case of this work NSGA-II.

We use CGP-NAS as a basis for this work, we extended several points, such as the representation and the number of blocks in the function set, as well as the addition of hyperameters to the search.

\begin{table}
\centering
\scriptsize
\caption{Summary of methods their objectives, datasets and encoding. }
\label{tab:SOTA}
\begin{tabular}{cccc}
\hline
Method &
  Objetives&
  Datasets &
  Encoding \\ \hline
\multicolumn{1}{|c|}{GeNet~\citep{Xie2017}} &
  \multicolumn{1}{c|}{Accuracy} &
  \multicolumn{1}{c|}{CIFAR-10, ImageNet} &
  \multicolumn{1}{c|}{Binary} \\ \hline
\multicolumn{1}{|c|}{CGP-CNN~\citep{Suganuma2017}} &
  \multicolumn{1}{c|}{Accuracy} &
  \multicolumn{1}{c|}{CIFAR-10, CIFAR-100} &
  \multicolumn{1}{c|}{CGP} \\ \hline
  \multicolumn{1}{|c|}{Torabi et al.~\citep{Torabi2022}} &
  \multicolumn{1}{c|}{Accuracy} &
  \multicolumn{1}{c|}{CIFAR-10, CIFAR-100} &
  \multicolumn{1}{c|}{CGP} \\ \hline

\multicolumn{1}{|c|}{AE-CNN~\citep{Sun2020}} &
  \multicolumn{1}{c|}{Accuracy} &
  \multicolumn{1}{c|}{CIFAR-10, CIFAR-100} &
  \multicolumn{1}{c|}{Block-Chained} \\ \hline
  
\multicolumn{1}{|c|}{DSGE-CNN~\citep{Lima2022}} &
  \multicolumn{1}{c|}{Accuracy} &
  \multicolumn{1}{c|}{CIFAR-10} &
  \multicolumn{1}{c|}{Grammar} \\ \hline
\multicolumn{1}{|c|}{Bakhshi et al.~\citep{Bakhshi2020}} &
  \multicolumn{1}{c|}{Accuracy} &
  \multicolumn{1}{c|}{\begin{tabular}[c]{@{}c@{}}CIFAR-10, CIFAR-100, \\ SVHN\end{tabular}} &
  \multicolumn{1}{c|}{Integer} \\ \hline
\multicolumn{1}{|c|}{NEMO~\citep{Kim2017}} &
  \multicolumn{1}{c|}{\begin{tabular}[c]{@{}c@{}}Accuracy, \\ Inference Speed\end{tabular}} &
  \multicolumn{1}{c|}{MNIST, CIFAR-10} &
  \multicolumn{1}{c|}{Integer} \\ \hline
\multicolumn{1}{|c|}{NSGA-Net~\citep{Lu}} &
  \multicolumn{1}{c|}{\begin{tabular}[c]{@{}c@{}}Accuracy,\\  MAdds\end{tabular}} &
  \multicolumn{1}{c|}{CIFAR-10, CIFAR-100} &
  \multicolumn{1}{c|}{Binary} \\ \hline
\multicolumn{1}{|c|}{NSGA-NetV1~\citep{Lu2020a}} &
  \multicolumn{1}{c|}{\begin{tabular}[c]{@{}c@{}}Accuracy, \\ FLOPs\end{tabular}} &
  \multicolumn{1}{c|}{\begin{tabular}[c]{@{}c@{}}CIFAR-10, CIFAR-100,\\ ImageNet\end{tabular}} &
  \multicolumn{1}{c|}{\begin{tabular}[c]{@{}c@{}}Block-Chained,\\ Graph\end{tabular}} \\ \hline
\multicolumn{1}{|c|}{NSGA-NetV2~\citep{Lu2021}} &
  \multicolumn{1}{c|}{\begin{tabular}[c]{@{}c@{}}Accuracy, \\ MAdds.\end{tabular}} &
  \multicolumn{1}{c|}{\begin{tabular}[c]{@{}c@{}}ImageNet,  CIFAR-10, \\ CIFAR-100\end{tabular}} &
  \multicolumn{1}{c|}{\begin{tabular}[c]{@{}c@{}}Block-Chained,\\ Integer\end{tabular}} \\ \hline
\multicolumn{1}{|c|}{MOCNN~\citep{Wang2020}} &
  \multicolumn{1}{c|}{\begin{tabular}[c]{@{}c@{}}Accuracy, \\ FLOPs\end{tabular}} &
  \multicolumn{1}{c|}{CIFAR-10} &
  \multicolumn{1}{c|}{Binary} \\ \hline
\multicolumn{1}{|c|}{MOGI-NET~\citep{Xue2021}} &
  \multicolumn{1}{c|}{\begin{tabular}[c]{@{}c@{}}Accuracy, \\ Parameters\end{tabular}} &
  \multicolumn{1}{c|}{CIFAR-10, CIFAR-100} &
  \multicolumn{1}{c|}{Binary} \\ \hline
\multicolumn{1}{|c|}{CGP-NAS~\citep{Garcia-Garcia2022}} &
  \multicolumn{1}{c|}{\begin{tabular}[c]{@{}c@{}}Accuracy, \\ MAdds\end{tabular}} &
  \multicolumn{1}{c|}{CIFAR-10, CIFAR-100} &
  \multicolumn{1}{c|}{CGP, Real} \\ \hline
  
\multicolumn{1}{|c|}{LF-MOGP~\citep{Liu2022}} &
  \multicolumn{1}{c|}{\begin{tabular}[c]{@{}c@{}}Accuracy, \\ parameters\end{tabular}} &
   \multicolumn{1}{c|}{\begin{tabular}[c]{@{}c@{}}CIFAR-10, CIFAR-100,\\ Fashion, MNIST variants\end{tabular}} &
  \multicolumn{1}{c|}{CGP} \\ \hline
  
\multicolumn{1}{|c|}{EvoApproxNAS~\citep{Pinos123}} &
  \multicolumn{1}{c|}{\begin{tabular}[c]{@{}c@{}}Accuracy, \\ parameters, \\ Energy\end{tabular}} &
  \multicolumn{1}{c|}{CIFAR-10,  SVHN} &
  \multicolumn{1}{c|}{CGP} \\ \hline
\end{tabular}
\end{table}

\subsection{Discussion}


Most of the methods discussed so far have either proposed new representations or adapted modifications to previous ones. Accordingly, crossover and mutation operators are also being developed. Having a variable-length encoding increased the NAS performance because the search space coverage increased. On the other hand, proposals that use information to estimate new solutions without training can significantly reduce the number of evaluations and have the advantage of using fewer GPU days. We can also note the limited use of bio-inspired methods such as PSO and the dominance of multiobjective evolutionary algorithms such as NSGA-II.


Our proposal implements a variable-length representation using CGP, that is also converted to a real-based encoding, thus avoiding any adaptation of the crossover and mutation operations. This speeds up CGP-NAS implementation through the ability of searching using different MOEAS as well as the possibility of having multiple types of crossover and mutation operators. Finally a summary of the methods presented here is presented in Table \ref{tab:SOTA}.





\label{pa}
\section{Continuous cartesian genetic programming based representation for multi-objective NAS}
The automated configuration of CNN architectures has been mostly driven by methods that aim to optimize a single objective, commonly, accuracy. While effective architectures can be obtained by optimizing model's performance, such an approach has a number of limitations, the most important is perhaps the propensity of solutions to overfitting because model complexity is not restricted. Likewise, model complexity is tied with efficiency, hence, low compexity models should be preferred. 
In this work, we propose the design of CNN architectures that  simultaneously maximize model's accuracy and minimize its complexity. The goal is to restrict the capacity of the model with the aim of obtaining accurate models based on architectures of moderated complexity. Compared to alternative solutions, ours is based on a CGP representation that allows us to operate in the real domain. This enables the usage of off-the-shell multi-optimization techniques. 

The approached problem  can be  formulated as one of  multi-objective optimization 
as follows: 
\begin{gather}
  \text{Minimize } \textbf{F}(x) =(f_1(x;w^*(x),f_2(x))^T\\
   \text{subject to: } w^*(x) \in \text{ argmin } \mathcal{L}(w;x)
   \label{genop}
\end{gather}  
 Where $f_1$ is an objective associated to the classification error of the CNN architecture as defined by parameters $w^*$ and $f_2$ is an objective associated to model complexity measured, in agreement with previous studies, with MAdds, as they express the total number of operations performed in each architecture. One should note that 
 MAdds are a guideline for certain implementation scenarios, such as under mobile settings, where the complexity must be less than or equal to 600 MAdds~\citep{Lu2020,Lu2021}.
 Please note that  in order  to obtain an estimate of classification error it is necessary to optimize the weights $w$ of the CNN architecture, so $x$ (solution) depends on this optimization where normally some training algorithm such as stochastic gradient descent (SDG) is used.

 Our approach to this problem relies on evolutionary algorithms, 
 with a set of individuals (a population of CNN architectures) being generated randomly, to be subsequently selected  according to their fitness. This subset of solutions goes through a process of crossover and mutation to evolve in new offspring. Finally, the new offspring and the current population are rearranged so that only the solutions with better fitness pass to the next generation. 

\begin{figure}[h]
    \centering
    \includegraphics[scale=0.7]{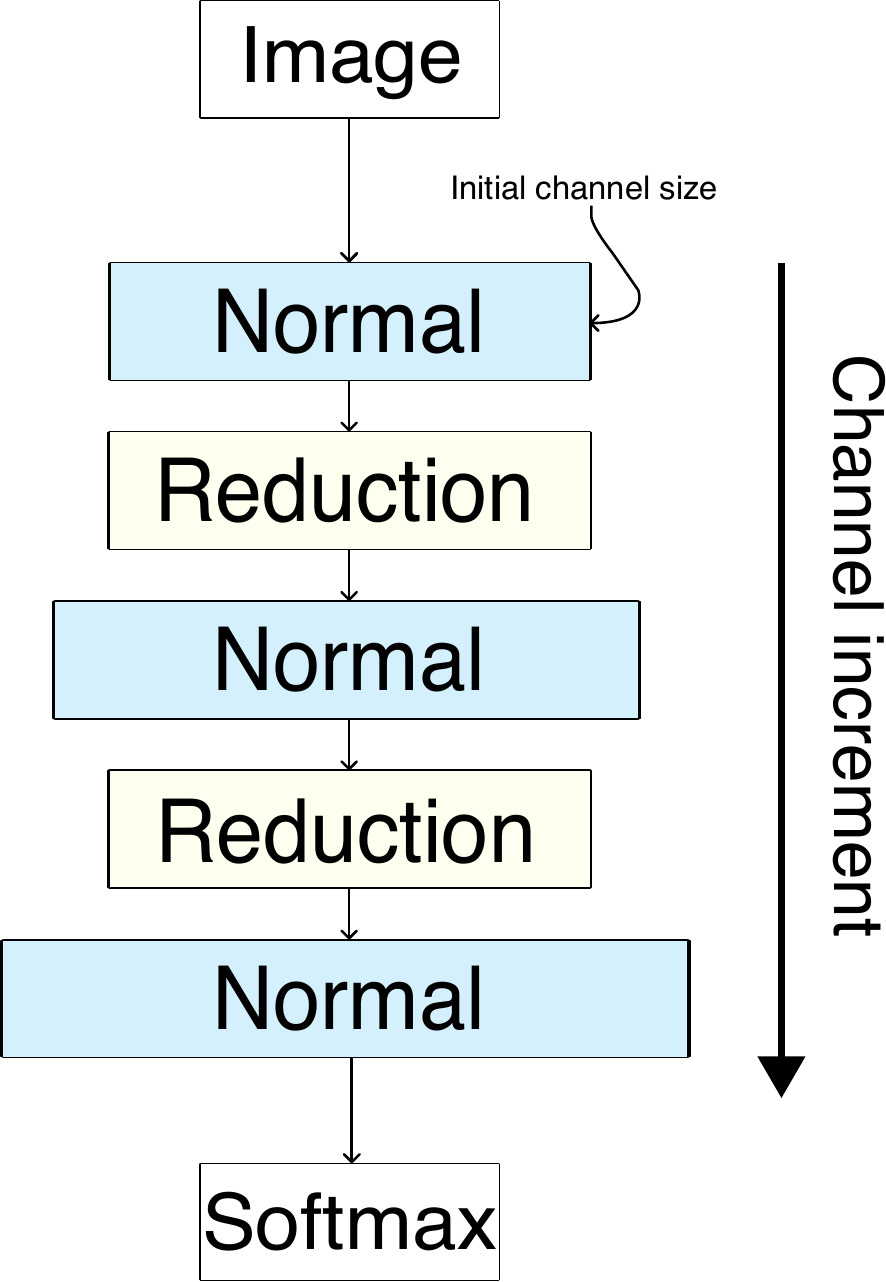}
    \caption{The general scheme of the representation  based on chained blocks; each block internally focuses on special operations; in each normal block, for example, convolution operations are performed; on the other hand, in the reduction blocks, methods such as pooling are applied to spatially reduce the feature maps. As the number of blocks increases, the number of channels will gradually increase.}
    \label{fig:normalandred}
\end{figure}

Two variants of our method are proposed. We first 
introduce a CGP-NASV1, a variant using a block-chained approach, see Figure~\ref{fig:normalandred}. In each ``Normal'' block, an internal CGP representation itself handles the connections and functions, in the ``Reduction'' blocks, a spatial reduction is applied, in our case, a pooling block is fixed. During the optimization process, this representation is re-encoded to a real-valued representation. Therefore, the full MOEAs searching potential within continuous domain is exploited.
After every generation, solutions are re-encoded to CGP and evaluated. 
In a second variant of CGP-NAS, we included the hyperparameters of the CNNs in the CGP representation, increasing the flexibility of model configurations that can be find, at the expense of increasing the search space. 
 
 The remainder of this section describes both variants in detail, we first describe the associated search space and then present the CGP representations for both variants, then the considered evolutionary algorithms are presented.



 \subsection{Search space}



CGP works on a set of functions which implicitly define the search space, see Table~\ref{tab:CgpFun}. We considered  $11$ functions for CGP-NASV1 , where we need to make explicit all of the possible combinations in the number of channels and kernels. 
 %
In CGP-NASV1, the representation is limited to three ``Normal'' blocks, where each block has a variable channel size. The first block handles functions with 32 or 64 output channels, the second block 64 or 128 and the third block 128 or 256, the larger the number of ``Normal'' blocks, the larger the number of output channels. 
Blocks with no variants are marked with  ``-''. The total number of functions-blocks in the function set is calculated as the possible combinations of output channels $C$ and Kernels $k$, which are in total $68$ functions-blocks.

In CGP-NASV2, the hyperparameters are encoded explicitly in the CGP representation, therefore it is not necessary to explicitly list all of the  combinations, hence the total number of functions are significantly reduced.
Adding the hyperparameters to the representation increases the size of the vector of decision variables and makes the problem slightly more complex, but on the other hand, it makes the representation more flexible as well as reduces the number of functions.


\begin{table}[h]
\footnotesize
\caption{Functions set with corresponding variations and arity}
\label{tab:CgpFun}
\begin{tabular}{ccccc}
\hline
Block type                         & Symbol                          & CGP-NASV1               & CGP-NASV2                 & Arity \\ \hline
\multicolumn{1}{c|}{ConvBlock} &
  \multicolumn{1}{c|}{$CB(C,k)$} &
  
  \multicolumn{1}{c|}{\begin{tabular}[c]{@{}c@{}}{$C\in\{32,64,128,256\}$}                       \\ $k\in$\{$1\times1,3\times3,5\times5,7\times7$\}\end{tabular}} &
  \multicolumn{1}{c|}{\begin{tabular}[c]{@{}c@{}}{$C\in\{32,64,128,256\}$}                       \\ $k\in$\{$3\times3,5\times5$\}\end{tabular}} &  1 \\
  
  \hline
  
\multicolumn{1}{c|}{ResBlock}      & \multicolumn{1}{c|}{$RB(C,k)$}  &  \multicolumn{1}{c|}{\begin{tabular}[c]{@{}c@{}}{$C\in\{32,64,128,256\}$}                       \\ $k\in$\{$3\times3,5\times5,7\times7$\}\end{tabular}} &
  \multicolumn{1}{c|}{\begin{tabular}[c]{@{}c@{}}{$C\in\{32,64,128,256\}$}                       \\ $k\in$\{$3\times3,5\times5$\}\end{tabular}}   & 1     \\
  
    \hline
\multicolumn{1}{c|}{Bottleneck}    & \multicolumn{1}{c|}{$BN(C,k)$}  &  \multicolumn{1}{c|}{\begin{tabular}[c]{@{}c@{}}{$C\in\{32,64,128,256\}$}                       \\ $k\in$\{$3\times3,5\times5,7\times7$\}\end{tabular}} &
  \multicolumn{1}{c|}{\begin{tabular}[c]{@{}c@{}}{$C\in\{32,64,128,256\}$}                       \\ $k\in$\{$3\times3,5\times5$\}\end{tabular}}   & 1     \\
      \hline
\multicolumn{1}{c|}{FusedMBconv}   & \multicolumn{1}{c|}{$FBC(C,k)$} & \multicolumn{1}{c|}{\begin{tabular}[c]{@{}c@{}}{$C\in\{32,64,128,256\}$}                       \\ $k\in$\{$3\times3$\}\end{tabular}} &
  \multicolumn{1}{c|}{\begin{tabular}[c]{@{}c@{}}{$C\in\{32,64,128,256\}$}                       \\ $k\in$\{$3\times3,5\times5$\}\end{tabular}} & 1     \\
  \hline
\multicolumn{1}{c|}{MBconv}        & \multicolumn{1}{c|}{$MBC(C,k)$} & \multicolumn{1}{c|}{Not used in CGP-NASV1} & 
  \multicolumn{1}{c|}{\begin{tabular}[c]{@{}c@{}}{$C\in\{32,64,128,256\}$}                       \\ $k\in$\{$3\times3,5\times5$\}\end{tabular}}   & 1     \\
  \hline
\multicolumn{1}{c|}{SepConv}       & \multicolumn{1}{c|}{$SC(C,k)$}  & \multicolumn{1}{c|}{\begin{tabular}[c]{@{}c@{}}{$C\in\{32,64,128,256\}$}                       \\ $k\in$\{$3\times3,5\times5,7\times7$\}\end{tabular}} &
  \multicolumn{1}{c|}{\begin{tabular}[c]{@{}c@{}}{$C\in\{32,64,128,256\}$}                       \\ $k\in$\{$3\times3,5\times5$\}\end{tabular}}    & 1     \\
\hline
\multicolumn{1}{c|}{DiConv}        & \multicolumn{1}{c|}{$DC(C,k)$}  & 
  \multicolumn{1}{c|}{\begin{tabular}[c]{@{}c@{}}{$C\in\{32,64,128,256\}$}                       \\ $k\in$\{$3\times3,5\times5$\}\end{tabular}} & 
  \multicolumn{1}{c|}{\begin{tabular}[c]{@{}c@{}}{$C\in\{32,64,128,256\}$}                       \\ $k\in$\{$3\times3,5\times5$\}\end{tabular}}  & 1     \\
  \hline
\multicolumn{1}{c|}{Indentity}     & \multicolumn{1}{c|}{$I$}        & \multicolumn{1}{c|}{-}  & \multicolumn{1}{c|}{-}    & 1     \\
  \hline

\multicolumn{1}{c|}{C1x7-7x1}      & \multicolumn{1}{c|}{$C17$}      & \multicolumn{1}{c|}{-}  & \multicolumn{1}{c|}{-}    & 1     \\
  \hline

\multicolumn{1}{c|}{Summation}     & \multicolumn{1}{c|}{Sum}        & \multicolumn{1}{c|}{-}  & \multicolumn{1}{c|}{-}    & 2     \\
  \hline

\multicolumn{1}{c|}{Concatenation} & \multicolumn{1}{c|}{Concat}     & \multicolumn{1}{c|}{-}  & \multicolumn{1}{c|}{-}    & 2     \\ \hline
\end{tabular}
\end{table}

In this work, we extended the number of blocks used in our previous work, adding more complex ones such as FusedMBConv\citep{TanEff} from EfficientNetV2, MBConv \citep{Sandler2018} from MobileNetV2, and Bottleneck blocks \citep{HeDEEP}, as well as blocks that have proven their effectiveness relatd work~\citep{Lu,Lu2020a} such as Dilated Convolution, Deep-wise separable Convolution and $1\times7$ then $7\times1$ Convolution. Moreover, the identity block is also included helping to generate more flexible architectures with no added computational cost.

\subsection{CGP-NASV1 solution representation}
\label{CGPV1}
\begin{figure}[h]
    \centering
    \includegraphics[scale=0.4]{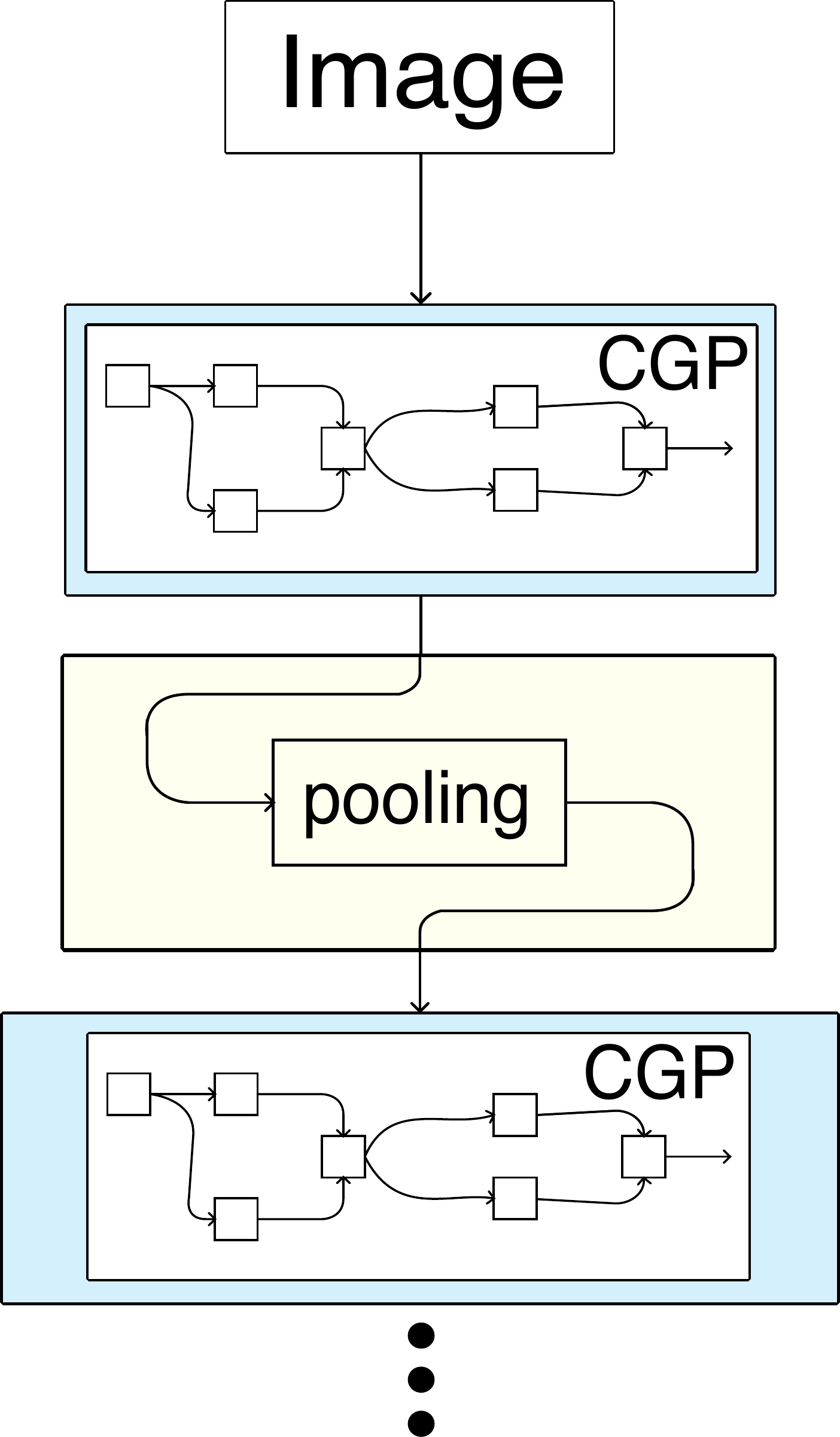}
    \caption{Block-chained representation}
    \label{fig:blockcha}
\end{figure}

CGP-NASV1 uses the block-chained encoding at the top level, it defines an template with some layers as shown in Figure \ref{fig:blockcha}. 
This representation connects blocks linearly, including those with specific tasks, in CGP-NASV1 the spatial reduction is performed by the pooling layers \citep{Lu2020a}.

At the top-level in the block-chained, CGP-NASV1 places a CGP within a ``Normal''
block while the reduction blocks implements a max pooling layer. In all pooling blocks, the pooling size is fixed as a $2\times2$ kernel and a stride of $2$. Figure \ref{fig:blockcha} shows an example, after the last block a global average pooling and a fully connected layer are added. Similar guidelines were found in the state of the art review \citep{Lu2020a,Pinos123}.

\begin{figure}[h]
    \centering
    \includegraphics[scale=0.43]{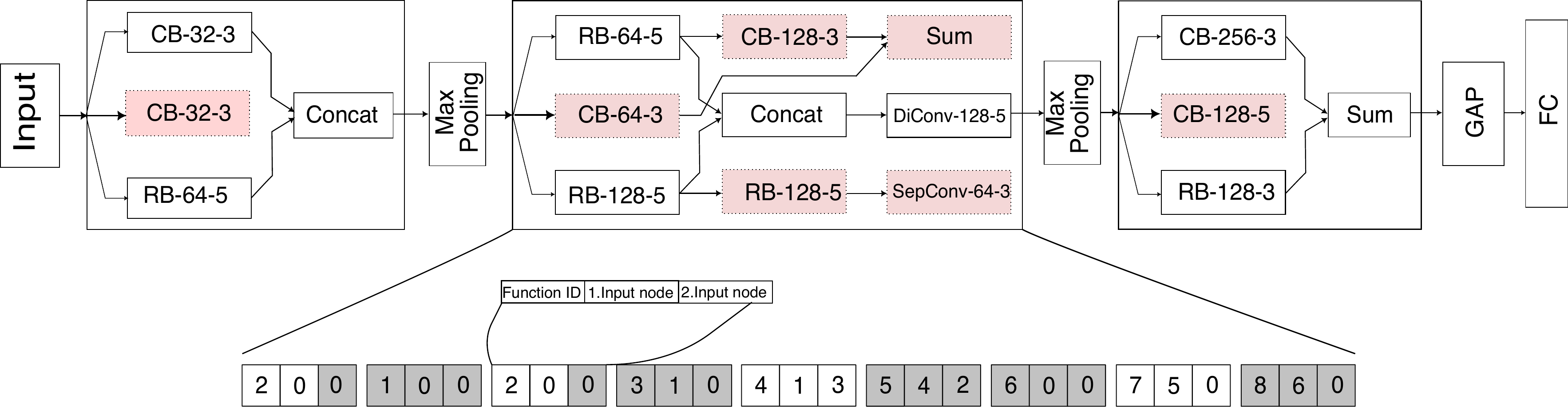}
    \caption{CGP-NASV1 using the block-chained representation}
    \label{fig:CCCGPNASV0}
\end{figure}

Figure \ref{fig:CCCGPNASV0} shows the final representation as an integer-based vector, where each square that encapsulates the function represents a "Normal" block with its own CGP. Each CGP is configured independently, for example, its own function set and size. In the figure, the red blocks contain the ID of a function.

%

We maintained the original idea from CGP-NAS of having two levels for representation~\citep{Garcia-Garcia2022}. In CGP-NASV1, each ``Normal'' block from the block-chained representation holds a CGP and it is represented as an integer vector.

At the second level of representation, the integer vector is encoded to a real-based vector. Equation \ref{funcenc} defines a range considering $func_k$ as the current function identifier and $func_{total}$ as the total number of functions in the function set. A uniform random number is generated to represent a function within this range in the real domain.

\begin{equation}
    rfunc_k\in\left[ \frac{func_k}{func_{total}}, \frac{func_k + 1}{func_{total}}\right]
    \label{funcenc}
\end{equation}

Equation \ref{inpuenc} defines a range for the function  inputs in the real domain to map connections for that node. This operation is applied to all connections. An input value ($nodeinput$) and its node number ($nodeTerm$) are used to calculate this range.

\begin{gather}
  rinput_j\in\left[ \frac{nodeinput_j}{nodeTerm}, \frac{nodeinput_j + 1}{nodeTerm}\right]
      \label{inpuenc} \\
 func_k=\lfloor gene_i \times func_{total}\rfloor
   \label{funcdec} \\
     input_j=\lfloor gene_i \times NodeTerm\rfloor
    \label{inputdec}
\end{gather}  
 
In order to decode solutions from the real to the integer domain, Equation \ref{funcdec} obtains the function identifier by multiplying the gene value by the total number of functions. Moreover, Equation \ref{inputdec} obtains the value of every connection by multiplying the gene value with the node number. The equations above are based on Clegg's work~\citep{Clegg2007}.





CGP-NASV1 represents CNN architectures in a divide and conquer approach. 
%
%
The use of a block-chained schema in synergy with CGP allows more control over the final architecture,


Our initial proposal CGP-NAS defined the size of the grid on which the complete neural architecture would be search. However, dealing with the search of an entire architecture at once can lead to very large and complex ones. This can be avoided, by dividing the overall architecture in more manageable stages. 


\subsection{CGP-NASV2 representation}
\label{CGPV2}

CGP-NASV2 proposal integrates the hyperparameters within the solutions representation of CGP-NASV1. Thus, they are also evolved and optimized during the search.
Changes in CGP-NASV2 are at the block-chained ``low level'' encoding. Therefore, there is  a reduction in the number of functions within the function set.
However, adding the hyperparameters to the search, necessarily increases the size of the integer vectors (and also the real-based vectors) to which the block-chained solutions are encoded.



In a deep neural architecture, there are hyperparameters such as the number of channels or filters and the kernel size. In CGP, these are explicitly included in the functions set if not considered within the solutions representation and in the evolutionary search. Therefore, a large function set must be defined in order to include all their possible combinations.
%
%
Moreover, a large function set directly increases the CGP grid size to hold them uniformly. 

In CGP-NASV2, we add the hyperparameters explicitly to the vector representing the CGP. This reduces the number of functions since only the standard functions remain and not their hyperparameters configuration.

\begin{figure}[h]
    \centering
    \includegraphics[scale=0.36]{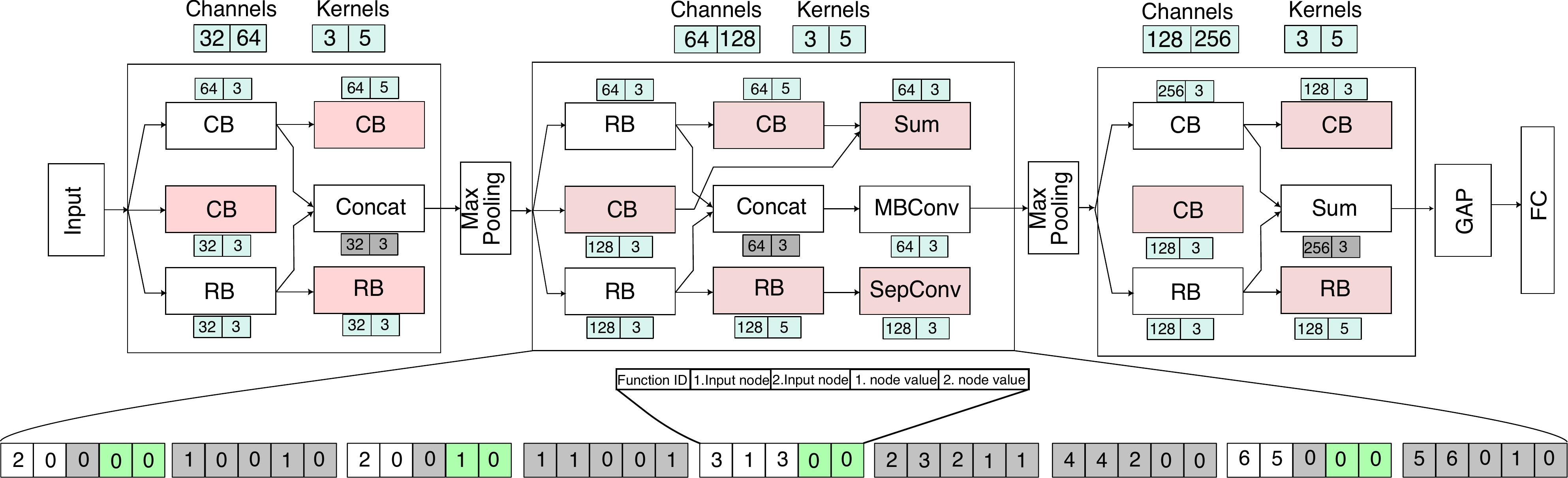}
    \caption{CGP-NASV2 block-chained representation with the hyperparameters directly encoded.}
    \label{fig:cgpW}
\end{figure}

In Figure \ref{fig:cgpW}, a possible solution in CGP-NASV2 is shown. The number of channels and kernel size are defined as parameters associated to each block-chained. The idea is to associate these parameters as weights to each CGP node. Green positions at the integer vector encoding in Figure \ref{fig:cgpW} represent the assigned hyperparameters within their corresponding CGP block.

\begin{figure}
     \centering
     \includegraphics[scale=0.35]{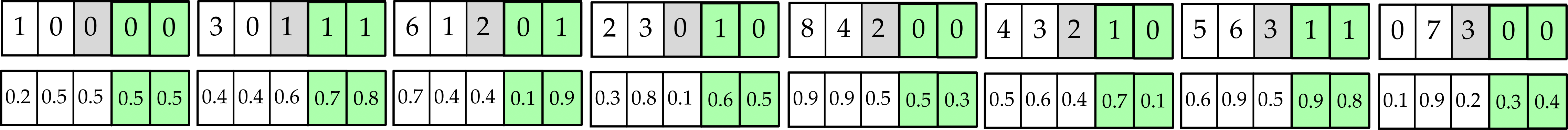}
     \caption{A CGP-NASV2 solution represented as an integer vector and its equivalent real-based vector with the added hyperparameters in the green positions. 
     }
     \label{fig:representationV1}
 
 \end{figure}

 Figure \ref{fig:representationV1} shows an example of a CGP-NASV2  solution representation as an integer vector with hyperparameters. 
 The entire vector is modified once the hyperparameters are explicitly considered by the CGP.
 The first three positions (white and gray positions) represent the id of the function and the input connections nodes. The green positions show two hyperparameters, one corresponding to the \#channels and the other to the \#kernels. 
 For example, in the integer vector representation from Figure \ref{fig:cgpW}, 0 in position 4 means that the value of the channel is 64 and 1 in position 5 means a kernel of 5x5. 

To convert the integer vector to the real-based vector, the same mechanism as the one used in CGP-NASV1 is applied adding only Equations \ref{funcencW} and \ref{inputdecW} to encode and decode the hyperparameters.

 \begin{gather}
  rHyp_k\in\left[ \frac{Hyp_k}{Hyp_{total}}, \frac{Hyp_k + 1}{Hyp_{total}}\right]
    \label{funcencW}\\
    Hyp_j=\lfloor gene_i \times Hyp_{total}\rfloor
    \label{inputdecW}
 \end{gather}
 

The  difference between both proposed variants is that in CGP-V1 the function set is restricted to explicit instantiations  of the function set. While in CGP-NASV2 this restriction does not hold, the algorithm itself has to figure out what functions what hyperparameters (filter size, channels) to use for building solutions. This provides more flexibility at the cost of increasing the search space. 
 

\subsection{Evolutionary algorithm}
\begin{figure}[ht]
    \centering
    \includegraphics[scale=0.7]{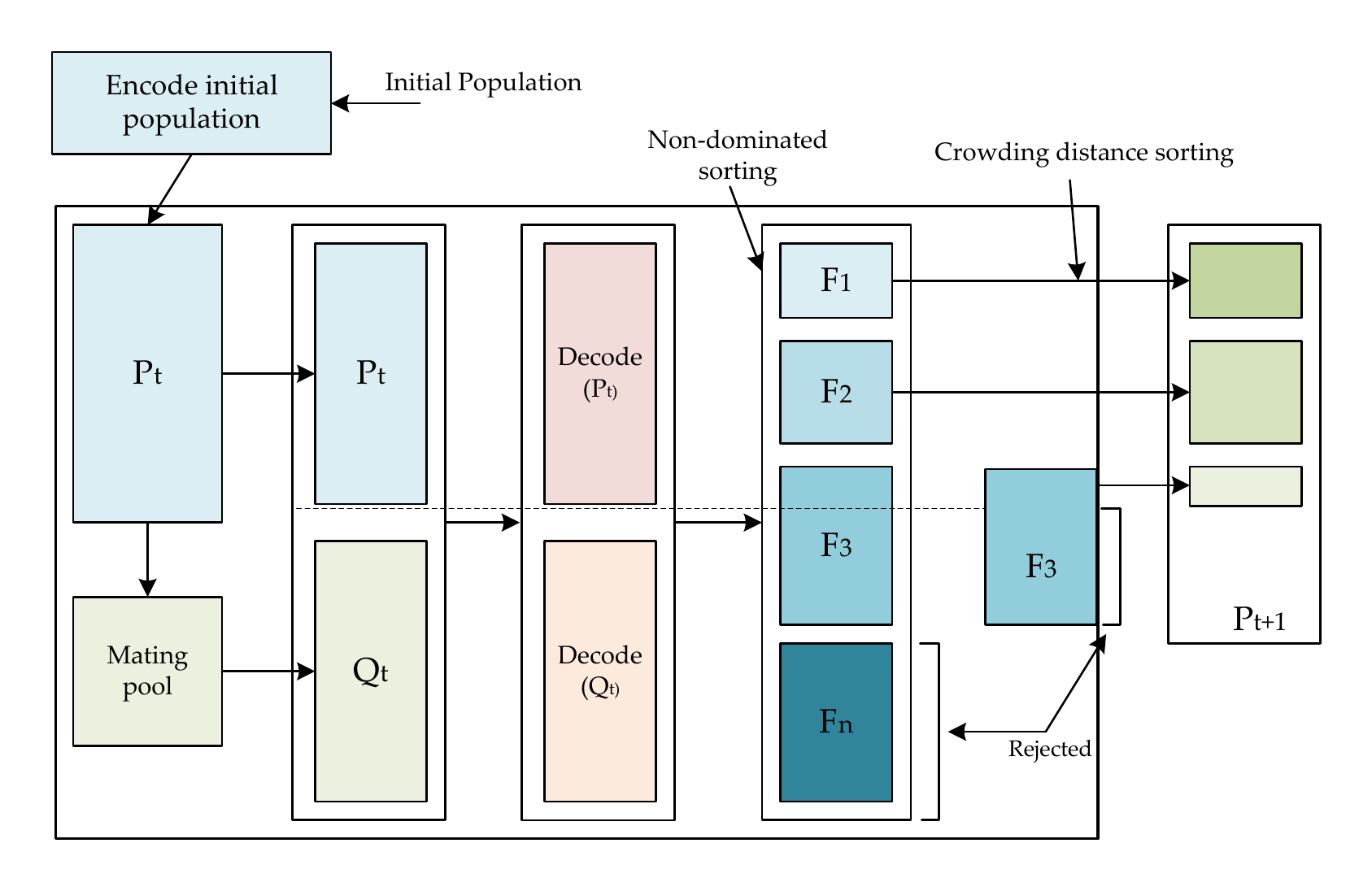}
    \caption{NSGA-II general schema with encoding and decoding steps. NSGA-II generates an initial population $Pt$, after mating an offspring population $Qt$ is obtained. $Qt$ is decoded for evaluation, and therefore for optimization considering the non-dominance criterion.}
    \label{fig:nsga-2}
\end{figure}

Working in a real-valued representations allows us to use any MOEA operating in such domain. Therefore, we initially considered the NSGA-II MOEA as optimizer, as this is one of the most used and effective MOEA.
Figure \ref{fig:nsga-2} shows a general schema of this widely known MOEA. Because, both CGP variants 
represent solutions in the continuous domain to perform the search, the full properties of NSGA-II are maintained. The only significant change is for the  the crossover and mutation operations. Both were adapted for CGP-NASV1 and CGP-NASV2 due to the use of block-chained representation.

\begin{figure}[ht]
    \centering
    \includegraphics[scale=0.25]{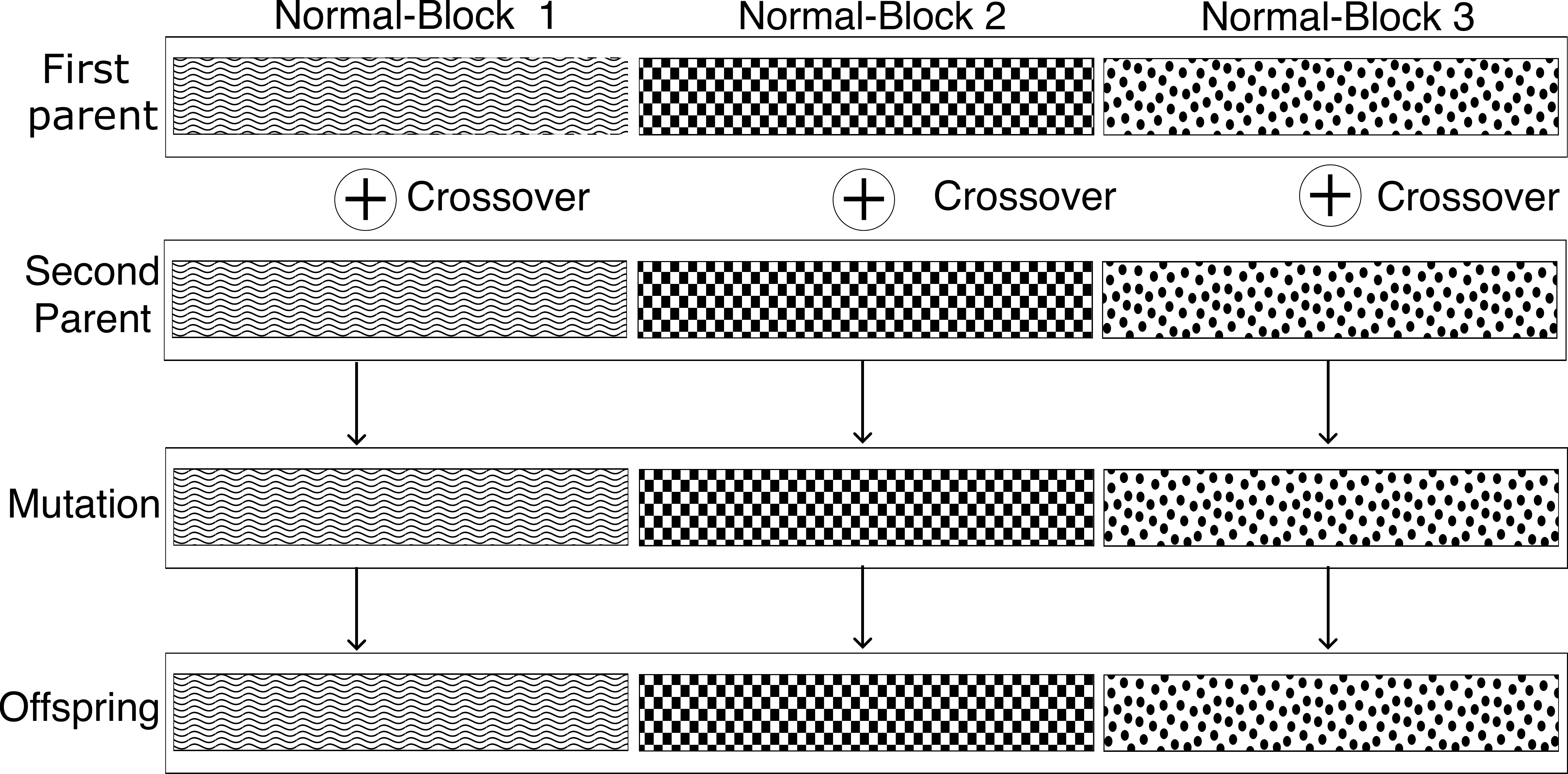}
    \caption{
    Each sub-vector represent a CGP ``normal'' block. Crossover applies independently between sub-vectors at the same overall position. Mutation works on the sub-vectors offspring.
    }
    
    \label{fig:matingpool}
\end{figure}

Each individual, represented as a real-based vector, is divided in three sub-vectors, each one encodes a CGP `Normal'' block. 
Figure~\ref{fig:matingpool} shows how the crossover and mutation operations are applied.
The crossover operation is performed between sub-vectors at the same overall position within the solution full vector. Also, the mutation operation is applied to the offspring of every sub-vector independently.  
It is important to emphasize that each operation is carried out independently at a sub-vector level. In the figure's example, three crossover and mutation operations are carried out for each sub-vector's offspring (pairs of CGP ``Normal'' blocks). These together represent the complete neural architecture to assess.
The crossover and mutation operators are defined by those commonly used in the NSGA-II; the simulated binary crossover (SBX) and the polynomial mutation (PM).

\subsection{ Fitness Function}
As previously mentioned in this paper we aim to simultaneously optimize two objectives with both variants of CGP-NAS, these are: 
 accuracy and the complexity as measured in MAdds. Where we want to maximize  accuracy and minimize the complexity.
The accuracy estimate is obtained by evaluating the validation set, which is a random subset of the training set, this way of obtaining the accuracy is called partial training performance estimation strategy~\citep{Elsken2019}.


We calculate the classification error after the accuracy using the following expression $1-Accuracy$, in order to minimize both objectives. 
A CNN architecture relies on the convolution operation which mainly performs multiplications and addition operations. Therefore, the complexity is calculated as the total number of multiplication-adds operations performed by the CNN. The complexity is calculated as the sum of multiplication and addition operations (MAdds) performed by the CNN architecture and is the second objective to minimize.

\label{es}

\section{Experimental framework}

In order to assess both proposed approaches, CGP-NASV1 and CGP-NASV2, first a direct performance comparison between the best evolved architecture based on the achieved accuracy as the performance metric is carried out. After,  a multiple-criteria decision analysis is performed to evaluate those evolved CNN models that better achieve a more balanced trade-off performance considering both objectives, accuracy and complexity. In this section, we present the experimental settings for both experiments as well as the considered  datasets. In the next section, we discuss the obtained results and thoroughly compared them against the original CGP-NAS~\citep{Garcia-Garcia2022} proposal and several state of the art approaches.
\begin{table}[h]
\small
\centering
\caption{CGP-NASV1 and CGP-NASV2 parameters}
\label{tab:confcgp}
\begin{tabular}{ccc}
\hline
\textbf{Parameters}                       & \textbf{CGP-NASV1}                           & \textbf{CGP-NASV2}       \\ \hline
\multicolumn{1}{c|}{Rows}        & \multicolumn{1}{c|}{5}              & 10             \\ \hline
\multicolumn{1}{c|}{Columns}     & \multicolumn{1}{c|}{25}             & 4              \\ \hline
\multicolumn{1}{c|}{Level-Back}  & \multicolumn{1}{c|}{1}             & 1              \\ \hline
\multicolumn{1}{c|}{Mutation probability} &
  \multicolumn{2}{c}{\begin{tabular}[c]{@{}c@{}}$Pm=0.3$
  \\ \end{tabular}} \\ \hline
\multicolumn{1}{c|}{Crossover probability} &
  \multicolumn{2}{c}{\begin{tabular}[c]{@{}c@{}}$Pc=0.9$, distribution index for\\  simulated binary crossover $Dsc=20$.\end{tabular}} \\ \hline
\multicolumn{1}{c|}{Population}  & \multicolumn{1}{c|}{24}             & 24             \\ \hline
\multicolumn{1}{c|}{Generations} & \multicolumn{1}{c|}{30} & 30  \\ \hline
\end{tabular}
\end{table}

Table \ref{tab:confcgp} shows CGP-NASV1 and CGP-NASV2 configuration in terms of the CGP and the evolutionary algorithm. The fundamental modification in CGP-NASV2 is having the hyperparameters directly represented within solutions and therefore a reduction in the grid size is also required due to the smaller functions set. In contrast, CGP-NASV1 requires a larger CGP grid size due to the bigger functions set associated to express all possible hyperparameters combination. 




\begin{figure}
    
     \begin{subfigure}[b]{0.5\textwidth}
         \centering
         \includegraphics[scale=0.5]{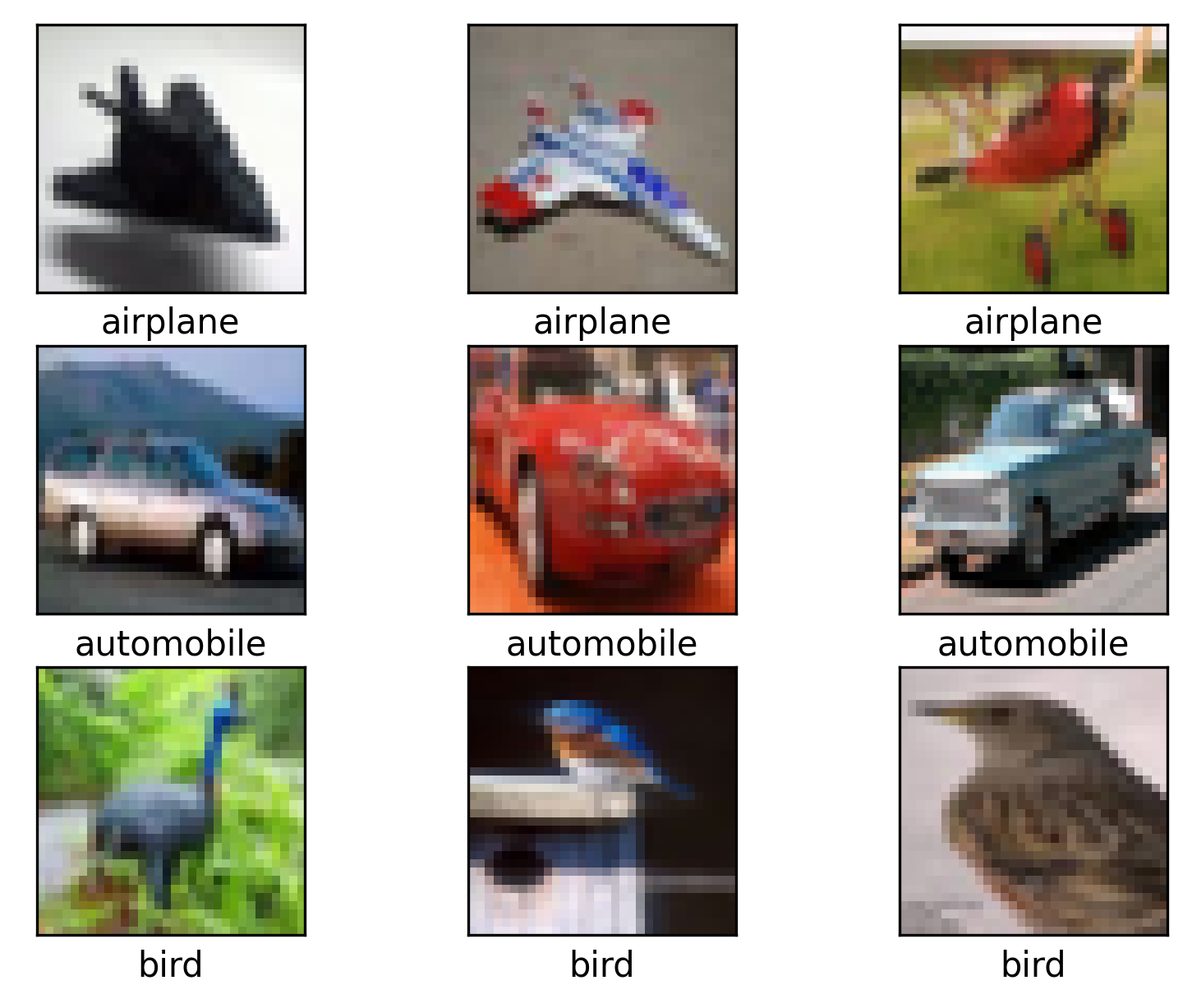}
         \caption{CIFAR-10  }
         \label{fig:c10D}
     \end{subfigure}
     \begin{subfigure}[b]{0.5\textwidth}
         \centering
         \includegraphics[scale=0.5]{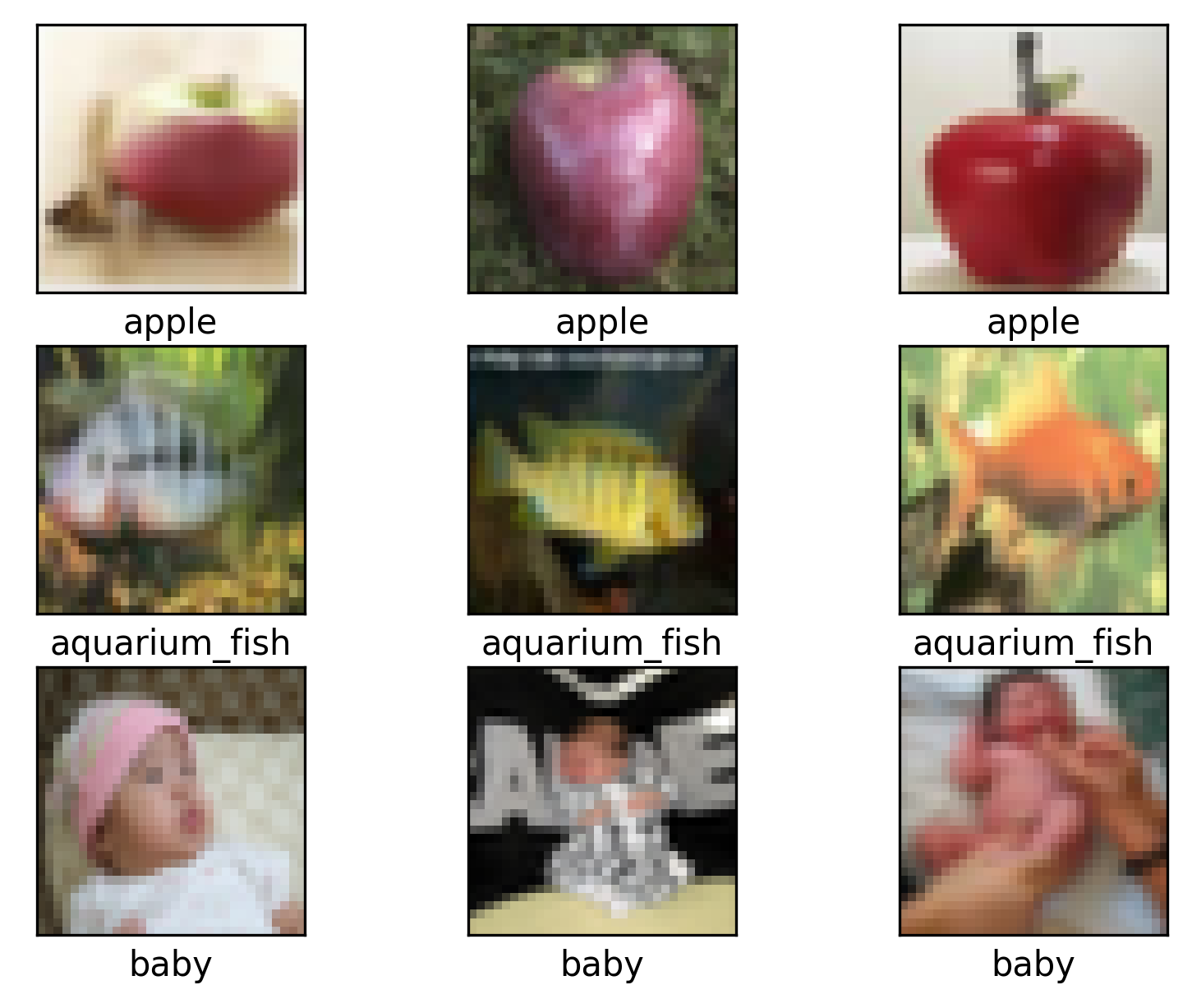}
         \caption{  CIFAR-100  }
         \label{fig:c100D}
     \end{subfigure}

        \caption{Examples from the CIFAR-10 and CIFAR-100 datasets, the image in each row belongs to the same class.}
        \label{fig:datasets}
\end{figure}

For evaluation of the proposed algorithmic approaches, two datasets for image classification were used: the CIFAR-100~\citep{Krizhevsky09learningmultiple} and the CIFAR-10~\citep{Krizhevsky09learningmultiple}. CIFAR-100 is a 100 classes image collection, while CIFAR-10 is a 10 classes, each dataset consists of 50,000 images for training and 10,000 for testing, both datasets contains $32 \times 32$ pixels color images. Figure~\ref{fig:datasets} shows several image samples of these datasets. 

For both, CGP-NASV1 and CGP-NASV2, during the evolutionary search, solutions were evaluated using a training set randomly divided in 40,000 images for training and 10,000 images for validation. 
For the final evaluation of the selected solutions of the Pareto front, the entire dataset is used: 50,000 images for training and 10,000 images for testing.

We use the following mechanisms for training the CNN architecture. The Stochastic Gradient Descent (SGD) was used, an as optimizer the cosine annealing learning rate schedule. 
The initial learning rate was set to $0.025$, while the momentum was set to $0.9$ and the weight decay to $0.0005$. A $128$ batch size was determined, in addition, $36$ training epochs were executed during evolutionary search.
Training and testing data were preprocessed with a $4$ pixel-mean subtraction padding on each size and randomly cropped with a $32 \times 32$ patch or its horizontally flipped image. 
Final solutions were re-trained for $600$ epochs and the cutout preprocessing technique was applied, the batch size was set to $96$.
Finally, an auxiliary head classifier~\cite{Lu} is used to improve the training process,  concatenated after the second reduction block, and the loss of this auxiliary head classifier is multiplied by a constant value of $0.4$ and added to the loss of the original architecture.
All experiments were executed on a Supercomputing Node with: 2 Intel Xeon E5-2650 v4 @ 2.20GHz processors, 8 Nvidia GTX 1080 Ti GPU cards and 128GB of RAM, running on Centos 7 OS.

\section{Results analysis}

We designed three experimental setups in order to evaluate the proposed algorithmic approaches. The difference between the first two experiments is the way the final solution is selected from the achieved Pareto front. In the first experiment, the final solution is determined by the smallest classification error, similar criterion is followed by most state of the art works. Therefore, we selected the evolved CNN presenting the lower classification error. After, this solution is executed $10$ times for a fair comparison between state of the art solutions, our original proposal CGP-NAS~\citep{Garcia-Garcia2022}, and the other algorithmic approaches herein discussed. 



In the second experiment, CGP-NASV1 and CGP-NASV2 were compared through a multiple-criteria decision-making (MCDM) method to find the solution with the best  Pareto front trade-off.

In the third experiment, CGP-NASV2 is assessed while implementing other MOEAs and searching operations .
In order to verify the representation developed, three different MOEAS were used: NSGA-II with the differential evolution crossover operator, MOEA/D, and SMS-EMOA.


%
%

Finally, we compare both proposed approaches versus the State of the Art works and discuss how the proposed solutions representation as well as the search mechanisms impact the final  CGP-NASV1 and CGP-NASV2 performances.

\subsection{Effectiveness of searching for the hyperparameters}


In the first experiment, we empirically assessed the effectiveness of including the hyperparameters in the evolutionary search. Thus, we compared CGP-NASV1 versus CGP-NASV2 on the CIFAR-10 and CIFAR-100 datasets. The best evolved architectures in terms of accuracy from $10$ independent runs were evaluated. We also added CGP-NAS as a baseline for comparison because it uses the same mechanism for selecting solutions from the Pareto front.

\begin{table}[h]
\caption{Parameters and MAdds expressed in millions ($1 \times 10^{6}$),}
\label{tab:CGPN}
\tiny
\begin{tabular}{|c|ccc|ccc|}
\hline
 & \multicolumn{3}{c|}{CIFAR10}                                         & \multicolumn{3}{c|}{CIFAR100}                                        \\ \hline
 & \multicolumn{1}{c|}{Error} & \multicolumn{1}{c|}{Parameters} & MAdds & \multicolumn{1}{c|}{Error} & \multicolumn{1}{c|}{Parameters} & MAdds \\ \hline
CGP-NAS~\citep{Garcia-Garcia2022} &
  \multicolumn{1}{c|}{\begin{tabular}[c]{@{}c@{}}4.86\\ (5.42 ± 0.46)\end{tabular}} &
  \multicolumn{1}{c|}{\begin{tabular}[c]{@{}c@{}}1.40\\ (2.52 + 0.90)\end{tabular}} &
  \begin{tabular}[c]{@{}c@{}}388.71\\ (1167.13+ 477.11)\end{tabular} &
  \multicolumn{1}{c|}{\begin{tabular}[c]{@{}c@{}}24.23\\ (26.41± 1.41)\end{tabular}} &
  \multicolumn{1}{c|}{\begin{tabular}[c]{@{}c@{}}5.43\\ (5.89 + 2.75)\end{tabular}} &
  \begin{tabular}[c]{@{}c@{}}1581.93\\ ( 1229.11 + 782)\end{tabular} \\ \hline
\multicolumn{1}{|l|}{CGP-NASV1} &
  \multicolumn{1}{c|}{\begin{tabular}[c]{@{}c@{}}4.23\\ (4.73 ± 0.44)\end{tabular}} &
  \multicolumn{1}{c|}{\begin{tabular}[c]{@{}c@{}}8.47\\ (8.74 ± 3.27)\end{tabular}} &
  \begin{tabular}[c]{@{}c@{}}1255.93\\ ( 1161.09 ± 373.78)\end{tabular} &
  \multicolumn{1}{c|}{\begin{tabular}[c]{@{}c@{}}21.76\\ (24.20 ± 1.8)\end{tabular}} &
  \multicolumn{1}{c|}{\begin{tabular}[c]{@{}c@{}}3.6\\ (7.25 ± 3.24)\end{tabular}} &
  \begin{tabular}[c]{@{}c@{}}791.85\\ (792.02 ± 342.6)\end{tabular} \\ \hline
\multicolumn{1}{|l|}{CGP-NASV2} &
  \multicolumn{1}{c|}{\textbf{\begin{tabular}[c]{@{}c@{}}3.70\\ (4.07± 0.17)\end{tabular}}} &
  \multicolumn{1}{c|}{\begin{tabular}[c]{@{}c@{}}4.04\\ (5.82 ±2.70)\end{tabular}} &
  \begin{tabular}[c]{@{}c@{}}636.32\\ (818.61 ± 372.62)\end{tabular} &
  \multicolumn{1}{c|}{\textbf{\begin{tabular}[c]{@{}c@{}}20.63\\ (22.49 ± 1.04)\end{tabular}}} &
  \multicolumn{1}{c|}{\begin{tabular}[c]{@{}c@{}}5.9\\ (6.50 ± 1.7)\end{tabular}} &
  \begin{tabular}[c]{@{}c@{}}827\\ (850.74± 476.08)\end{tabular} \\ \hline
\end{tabular}
\end{table}

Table \ref{tab:CGPN} shows the performance effect of evolving the hyperparameters. CGP-NASV2 reduces the error by half percentile point in comparison to CGP-NASV1 and in more than one percentile point with respect to the baseline approach CGP-NAS. This reduction is also observed for the standard deviation corresponding to the $10$ experimental samples, demonstrating CGP-NASV2 is robust in comparison to the other approaches.


We can also notice that the complexity measured in MAdds achieved by CGP-NASV2 is lower than that obtained by CGP-NASV1 for the CIFAR-10 dataset. On the other hand, both have similar values in MAdds terms for the CIFAR-100 dataset. 
It is worth mentioned that, the number of parameters is not one of objectives to optimize, therefore a significant variation among results is observed.
Selecting the best evolved architecture by accuracy, easily shows that CGP-NASV2 achieves the best results in terms of the classification error. On contrast, the complexity objective is negatively affected by having also a higher standard deviation.


\subsection{Best trade-off solution via multiple-criteria decision analysis}


CGP-NASV1 and CGP-NASV2 are multi-objective proposals, therefore selecting the best solution in terms of accuracy from the Pareto front is not an efficient criterion. When evolving neural architectures as a multi-objective optimization problem, the best solution would represent a trade-off between objectives. Therefore, in the proposed algorithmic approaches, a trade-off between the classification error and the complexity of the architecture in terms of Madds. 
The Multi-criteria decision making (MCDM) was used to select the best solution from the Pareto front, in particular, the \textit{ Knee and Boundary Selection} method~\citep{FernandesJr.2021} has been explored for the empirical analysis.

\begin{figure}[h]
    \centering
    \includegraphics[scale=0.9]{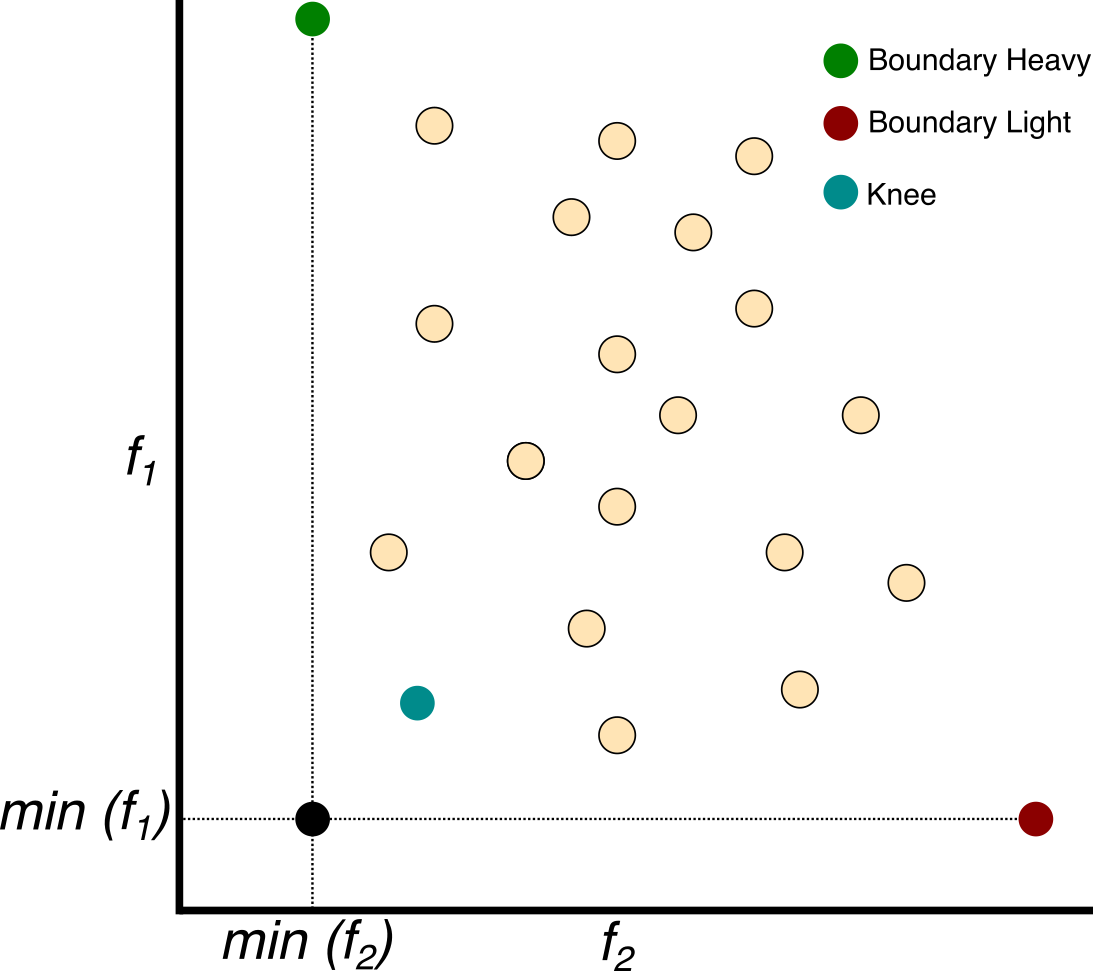}
    \caption{Knee and boundary selection
method. Measuring the distance between the intersections of the two best extreme solutions ensures the best trade-off solution is obtained.}
    \label{fig:kneeandb}
\end{figure}

This method obtains two solutions, called boundary heavy and boundary light, corresponding to the solution with the lowest value per objective. After, it calculates the solution closest to the intersection of both solutions (heavy and light), the obtained solution is called the Knee and represents the solution with the best trade-off. Figure \ref{fig:kneeandb} shows how the knee and boundary selection works.
%
%
%
%
Table \ref{tab:KNEENAS} shows the results achieved after applying the MCDM analysis to compare CGP-NASV1 and CGP-NASV2 in both CIFAR-10 and CIFAR-100 datasets.  
\begin{table}[h]
\caption{Trade-off knee solutions from the Pareto front, parameters and MAdds are expressed in millions ($1 \times 10^{6}$)}
\label{tab:KNEENAS}
\tiny
\begin{tabular}{|l|ccc|ccc|}
\hline
\multicolumn{1}{|c|}{} & \multicolumn{3}{c|}{CIFAR10}                                         & \multicolumn{3}{c|}{CIFAR100}                                        \\ \hline
\multicolumn{1}{|c|}{} & \multicolumn{1}{c|}{Error} & \multicolumn{1}{c|}{Parameters} & MAdds & \multicolumn{1}{c|}{Error} & \multicolumn{1}{c|}{Parameters} & MAdds \\ \hline
CGP-NASV1 &
  \multicolumn{1}{c|}{\begin{tabular}[c]{@{}c@{}}5.66\\ (7.33± 1.55)\end{tabular}} &
  \multicolumn{1}{c|}{\begin{tabular}[c]{@{}c@{}}0.41\\ (8.74 ± 0.13)\end{tabular}} &
  \begin{tabular}[c]{@{}c@{}}70.13\\ ( 52.01± 21.09)\end{tabular} &
  \multicolumn{1}{c|}{\begin{tabular}[c]{@{}c@{}}31.47\\ (33.34± 1.6)\end{tabular}} &
  \multicolumn{1}{c|}{\begin{tabular}[c]{@{}c@{}}1.06\\ (0.82 ± 1.12)\end{tabular}} &
  \begin{tabular}[c]{@{}c@{}}45.70\\ (30.81 ± 9.7)\end{tabular} \\ \hline
CGP-NASV2 &
  \multicolumn{1}{c|}{\textbf{\begin{tabular}[c]{@{}c@{}}4.85\\ (5.59± 0.5)\end{tabular}}} &
  \multicolumn{1}{c|}{\begin{tabular}[c]{@{}c@{}}0.78\\ (0.71 ±0.31)\end{tabular}} &
  \begin{tabular}[c]{@{}c@{}}53.99\\ (79.44± 31.96)\end{tabular} &
  \multicolumn{1}{c|}{\textbf{\begin{tabular}[c]{@{}c@{}}23.57\\ (28.23 ± 2.20)\end{tabular}}} &
  \multicolumn{1}{c|}{\begin{tabular}[c]{@{}c@{}}0.49\\ (0.53 ± 0.13)\end{tabular}} &
  \begin{tabular}[c]{@{}c@{}}66.66\\ (55.66± 19.14)\end{tabular} \\ \hline
\end{tabular}
\end{table}



The differences that we found with this selection method are mostly solutions with a lower number of parameters as well as MAdds. Although the classification error increased in comparison to that obtained by selecting the solution with the lowest classification error, the difference is relatively minor. Averaging 2\% and 5\% increase on CIFAR-10 and  CIFAR-100 respectively. However, the reduction in terms of MAdds from 818.61M to 79.44M for CIFAR-10 and from 850.74M to 55.66M for CIFAR-100 represents a substantial decrease in model complexity. Highlighting the ability to find solutions that balance both objectives. Additionally, our results indicate that incorporating the hyperparameters within the search process is beneficial, particularly in terms of reducing the classification error.


\subsection{Comparison versus the State of the Art}

Tables \ref{tab:Presults10} and \ref{tab:Presults100} present a detailed comparison between the State of the Art works and the proposed algorithmic approaches CGP-NASV1 and CGP-NASV2 on the CIFAR-10 and CIFAR-100 datasets. A total of $4$ human designs, $5$ NAS single-objective and $7$ mutli-objective proposals are considered for comparison plus the original proposal CGP-NAS~\cite{Garcia-Garcia2022} is also included in both tables. 

From previous empirical assessments, it was determined that CGP-NASV2 provided the best overall results when compared to CGP-NASV1 and their baseline CGP-NAS. It was also analyzed that selecting the best solution from the Pareto front in terms of classification error would negatively affect the resulting evolved architecture in its complexity. Thus, a method to select trading-off solutions for both conflicting objects known as the knee and boundary methods allowed the selection of an evolved architecture with more balanced performance metrics. In evolutionary algorithms particularly in the multi-objective optimization domain, Pareto front solutions represent a trade-off between conflicting objectives.   


 

\begin{table}
\centering
\caption{
Comparison on CIFAR-10 dataset: Classification error rate, the number of parameters and Multiply-adds (MAdds) are expressed in millions ($1 \times 10^{6}$), GPU-days and GPU Hardware. 
}
\label{tab:Presults10}
\resizebox{14cm}{!}{  
\begin{tabular}{cccccc}
\hline
Model &
  \begin{tabular}[c]{@{}c@{}}Error\\ rate \% \end{tabular} &
Params &
MAdds &GPU-Days& GPU hardware \\ \hline
\multicolumn{5}{c}{\textbf{Human Design}}\\
\hline

DenseNet $(k=12)$     (Huang et al., 2017)~\citep{huang2017densely} \nocite{huang2017densely}   & 5.24 & 1.0   & -    & -                                                           \\\hline
ResNet $(depth=101)$   (He et al., 2016)~\citep{he2016deep} & 6.43 & 1.7   & -    & -                                                           \\\hline
ResNet $(depth=1202)$ (He et al., 2016)~\citep{he2016deep} & 7.93 & 10.2  & -    & -                                                           \\\hline
VGG         (Simonyan et al., 2014)~\citep{simonyan2014very}          & 6.66 & 20.04 & -    & -                                                           \\\hline
\multicolumn{5}{c}{\textbf{Single Objective Approaches}}   \\  \hline

CGP-CNN(ConvSet) (Suganuma et al., 2020)~\citep{Suganuma2020} &  5.92 &  1.50 &  - &  8 & Nvidia 1080Ti \\\hline
CGP-CNN(ResSet) (Suganuma et al., 2020)~\citep{Suganuma2020}&  5.01 &  3.52 &  - &  14.7& Nvidia 1080Ti \\\hline
Large-Scale Evolution (Real et al., 2017)~\citep{Real2017} & 5.4  & 5.4  & -    & 2750& - \\\hline

AE-CNN (Sun et al., 2020)~\citep{Sun2020} & 4.3 & 2.0   & -    & 27& Nvidia 1080 Ti
\\\hline
Genetic-CNN (Xie et al., 2017)~\citep{Xie2017} & 7.1 & -   & -    & 17& -

\\\hline
 (Torabi et al., 2022)~\citep{Torabi2022}&
  5.69  &
  1.96 &
  - &
  - & NVIDIA Tesla k80
 \\\hline


\multicolumn{5}{c}{\textbf{Multi-Objective Approaches}}  \\
\hline
 NSGANet (Lu et al., 2020)~\citep{Lu2020}&
  3.85 &
  3.3 &
  1290 &
  8 & Nvidia 1080 Ti \\\hline 
NSGANetV1 (Lu et al., 2020)~\citep{Lu2020}&
  4.67 &
  0.2 &
  - &
  27 & Nvidia 2080 Ti \\\hline
 MOCNN (Wang et al., 2021)~\citep{Wang2020} &
  4.49 &
  - &
  - &
  24 & Nvidia 1080 Ti 
  
 \\\hline 
  
MOGIG-Net (Xue et al., 2021)~\citep{Xue2021} &
  4.67 &
  0.2 &
  - &
  14&- 
  
 \\\hline

EEEA-Net (Termritthikun et al., 2021)~\citep{Termritthikun2021}&
  2.46  &
  3.6 &
  - &
  0.52 & Nvidia RTX 2080 Ti 
 \\\hline

EvoApproxNAS (Pinos et al., 2022)~\citep{Pinos123}&
  6.80  &
  1.11 &
  458.2 &
  8.8 & NVIDIA Tesla V100-SXM2
 \\\hline

 LF-MOGP(Liu et al., 2022)~\citep{Liu2022}&
  4.13  &
  1.07 &
  - &
  10 & NVIDIA GeForce 3090
 \\\hline

\textbf{CGP-NAS} (Garcia-Garcia et al., 2022)~\citep{Garcia-Garcia2022} & \begin{tabular}[c]{@{}c@{}}\textbf{4.86}\\ \textbf{(5.42 $\pm$ 0.46)}\end{tabular} &\textbf{ 1.40} & \textbf{388.71} & \textbf{1.4}& \textbf{Nvidia Titan X} 
\\ \hline




\textbf{CGP-NASV2-Best solution} & \begin{tabular}[c]{@{}c@{}}\textbf{3.70}\\ \textbf{(4.07$\pm$ 0.17)}\end{tabular} &\begin{tabular}[c]{@{}c@{}}\textbf{4.04}\\ \textbf{(5.82 $\pm$ 2.70)}\end{tabular} & \begin{tabular}[c]{@{}c@{}}\textbf{636.32}\\ \textbf{(818.61 $\pm$ 372.62)}\end{tabular}&  11.54&\textbf{Nvidia 1080Ti} \\    
\hline

\textbf{CGP-NASV2-Knee solution} & \begin{tabular}[c]{@{}c@{}}\textbf{4.85}\\ \textbf{(5.59 $\pm$ 0.5)}\end{tabular} &\begin{tabular}[c]{@{}c@{}}\textbf{0.78}\\ \textbf{(0.71 $\pm$ 0.31)}\end{tabular} & \begin{tabular}[c]{@{}c@{}}\textbf{53.99}\\ \textbf{(79.44 $\pm$ 31.96)}\end{tabular}&  11.54&\textbf{Nvidia 1080Ti} \\    

\hline


\end{tabular}
}

\end{table}


\begin{table}[h]
\centering
\caption{
Comparison on CIFAR-100 dataset: Classification error rate, the number of parameters and Multiply-adds (MAdds) are expressed in millions ($1 \times 10^{6}$), GPU-days and GPU Hardware. 
}
\label{tab:Presults100}
\resizebox{14cm}{!}{  
\begin{tabular}{cccccc}
\hline
Model &
  \begin{tabular}[c]{@{}c@{}}Error\\ rate \% \end{tabular} &
Params &
MAdds &GPU-Days& GPU hardware \\ \hline
\multicolumn{5}{c}{\textbf{Human Design}}\\
\hline

DenseNet $(k=12)$     (Huang et al., 2017)~\citep{huang2017densely} \nocite{huang2017densely}   & 24.42 & 1.0   & -    & -                                                           \\\hline
ResNet $(depth=101)$   (He et al., 2016)~\citep{he2016deep} & 25.16 & 1.7   & -    & -                                                           \\\hline
ResNet $(depth=1202)$ (He et al., 2016)~\citep{he2016deep} & 27.82 & 10.2  & -    & -                                                           \\\hline
VGG         (Simonyan et al., 2014)~\citep{simonyan2014very}          & 28.05 & 20.04 & -    & -                                                           \\\hline
\multicolumn{5}{c}{\textbf{Single Objective Approaches}}   \\  \hline

CGP-CNN(ConvSet) (Suganuma et al., 2020)~\citep{Suganuma2020} &  26.7 &  2.04 &  - &  13 & Nvidia 1080Ti \\\hline
CGP-CNN(ResSet) (Suganuma et al., 2020)~\citep{Suganuma2020}&  25.1 &  3.43 &  - &  10.9& Nvidia 1080Ti \\\hline
Large-Scale Evolution (Real et al., 2017)~\citep{Real2017} & 23.0  & 40.4  & -    & 2750& - \\\hline

AE-CNN (Sun et al., 2020)~\citep{Sun2020} & 20.85 & 5.4   & -    & 36& Nvidia 1080 Ti
\\\hline
Genetic-CNN (Xie et al., 2017)~\citep{Xie2017} & 29.03 & -   & -    & 17& -
\\\hline

 (Torabi et al., 2022)~\citep{Torabi2022}&
  26.03  &
  2.56 &
  - &
 - & NVIDIA Tesla V100-SXM2
 \\\hline

\multicolumn{5}{c}{\textbf{Multi-Objective Approaches}}  \\
\hline
  
NSGANetV1 (Lu et al., 2020)~\citep{Lu2020}&
  25.17 &
  0.2 &
  1290 &
  27 & Nvidia 2080 Ti \\\hline

MOGIG-Net (Xue et al., 2021)~\citep{Xue2021} &
  24.71 &
  0.7 &
  - &
  14&- 
  
 \\\hline

EEEA-Net (Termritthikun et al., 2021)~\citep{Termritthikun2021}&
  15.02  &
  3.6 &
  - &
  0.52 & Nvidia RTX 2080 Ti 
 \\\hline

 LF-MOGP(Liu et al., 2022)~\citep{Liu2022}&
  26.37  &
  4.12 &
  - &
  13 & NVIDIA GeForce 3090
 \\\hline

\textbf{CGP-NAS(Garcia-Garcia et al., 2022)~\citep{Garcia-Garcia2022}} & \begin{tabular}[c]{@{}c@{}}\textbf{24.23}\\ \textbf{(26.41 $\pm$ 1.41)}\end{tabular} &\textbf{ 5.43} & \textbf{1581} & \textbf{2.1}& \textbf{Nvidia Titan X} 
\\ \hline

\textbf{CGP-NASV2 - Best Solution} & \begin{tabular}[c]{@{}c@{}}\textbf{20.63}\\ \textbf{(22.49 $\pm$ 1.04)}\end{tabular} &\begin{tabular}[c]{@{}c@{}}\textbf{5.9}\\ \textbf{(6.50$\pm$ 1.7)}\end{tabular} & \begin{tabular}[c]{@{}c@{}}\textbf{827}\\ \textbf{(850.74 $\pm$ 476.08)}\end{tabular}&  11.28&\textbf{Nvidia 1080Ti} \\    
\hline
\textbf{CGP-NASV2 - Knee solution} & \begin{tabular}[c]{@{}c@{}}\textbf{23.57}\\ \textbf{(28.43 $\pm$ 2.20)}\end{tabular} &\begin{tabular}[c]{@{}c@{}}\textbf{0.49}\\ \textbf{(0.53 $\pm$ 0.13)}\end{tabular} & \begin{tabular}[c]{@{}c@{}}\textbf{66.66}\\ \textbf{(55.66 $\pm$ 19.14)}\end{tabular}&  11.28&\textbf{Nvidia 1080Ti} \\  
\hline


\end{tabular}
}

\end{table}

\begin{figure}[h]
    
     \begin{subfigure}[b]{0.5\textwidth}
         \centering
         \includegraphics[width=\textwidth]{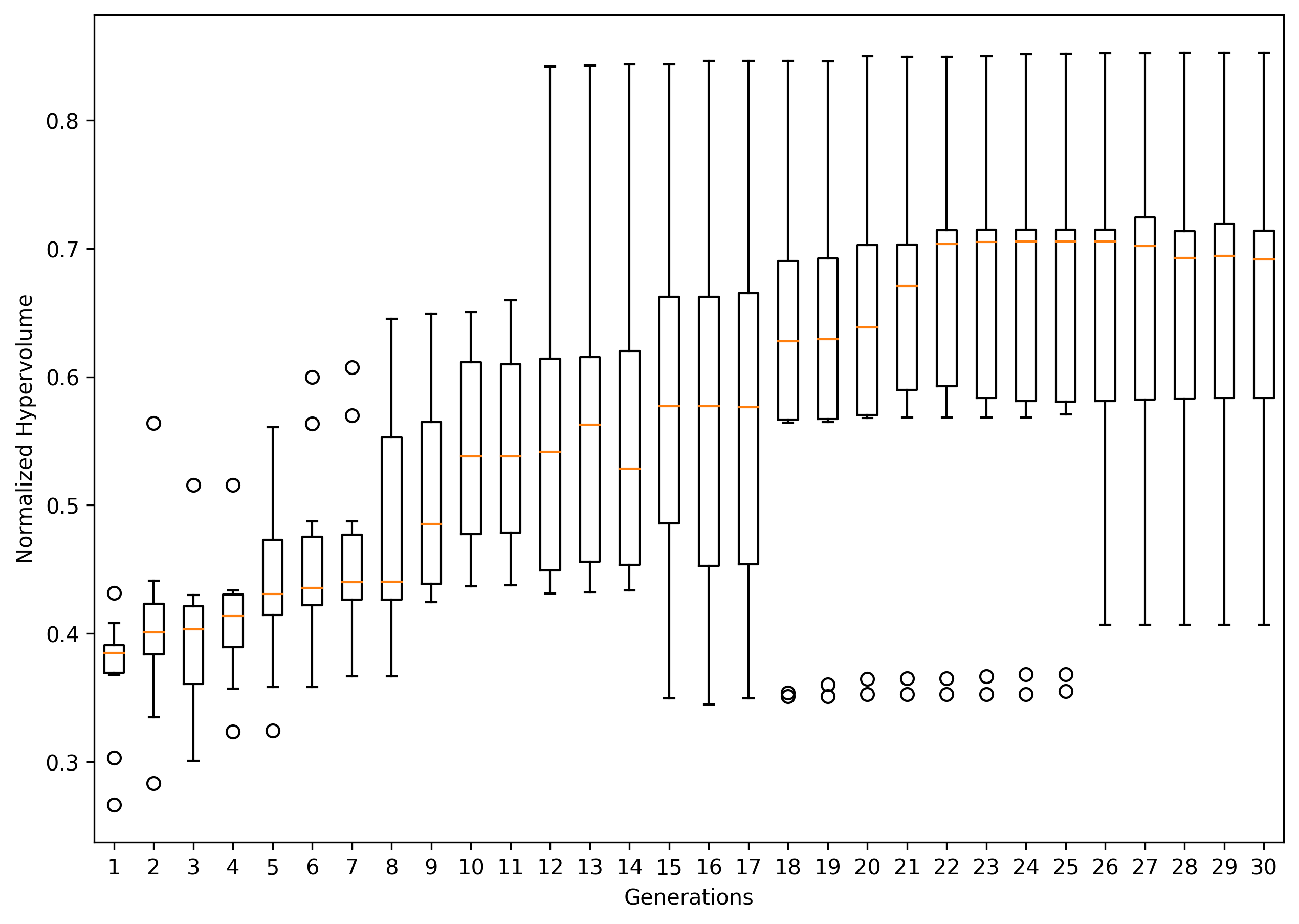}
         \caption{Hypervolume CGPNAS-V1 on CIFAR-10}
         \label{fig:c10hv1}
     \end{subfigure}
     \begin{subfigure}[b]{0.5\textwidth}
         \centering
         \includegraphics[width=\textwidth]{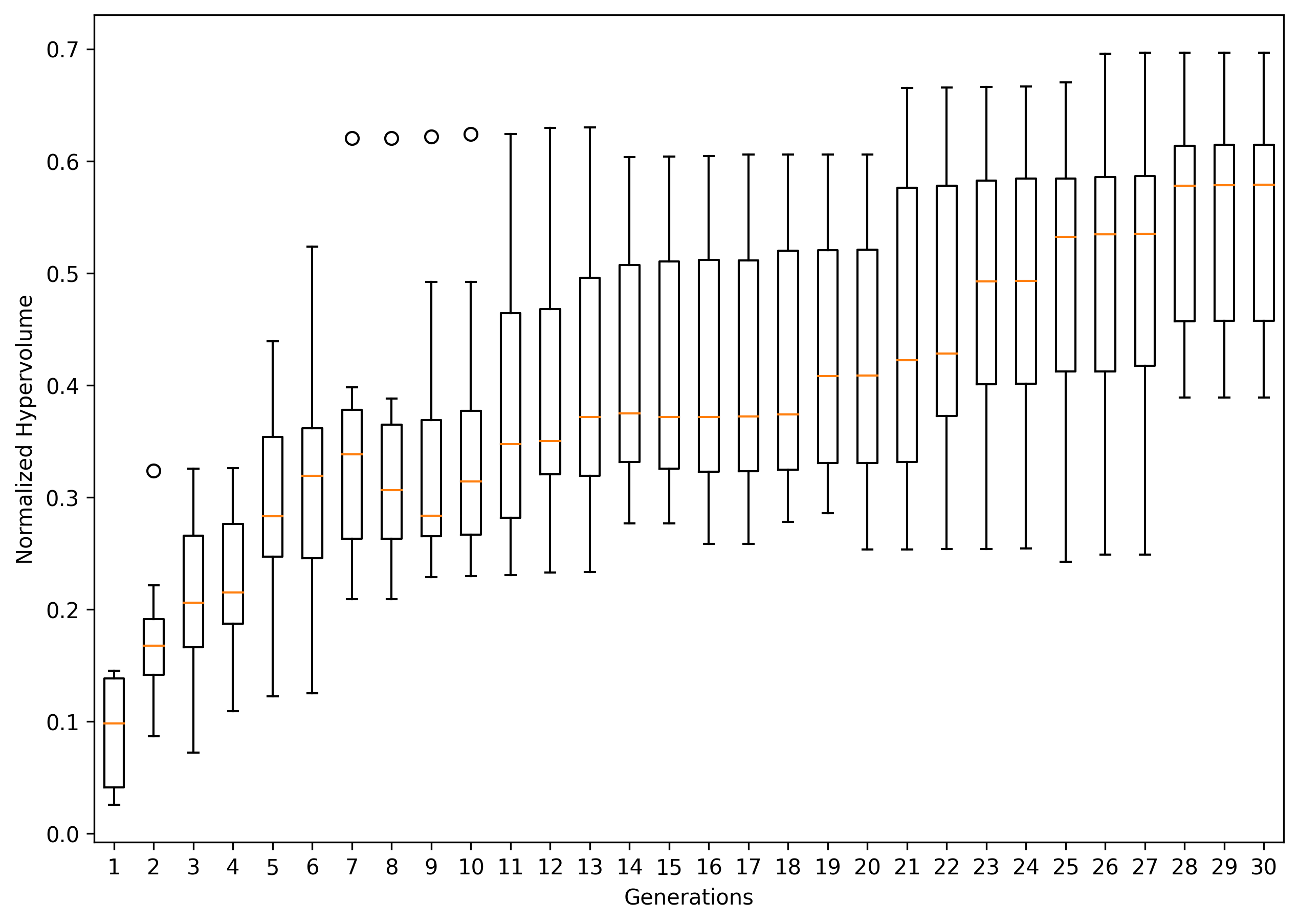}
         \caption{Hypervolume CGPNAS-V2 on CIFAR-10}
         \label{fig:c10hv2}
     \end{subfigure}
     \hfill
     \begin{subfigure}[b]{0.5\textwidth}
         \centering
         \includegraphics[width=\textwidth]{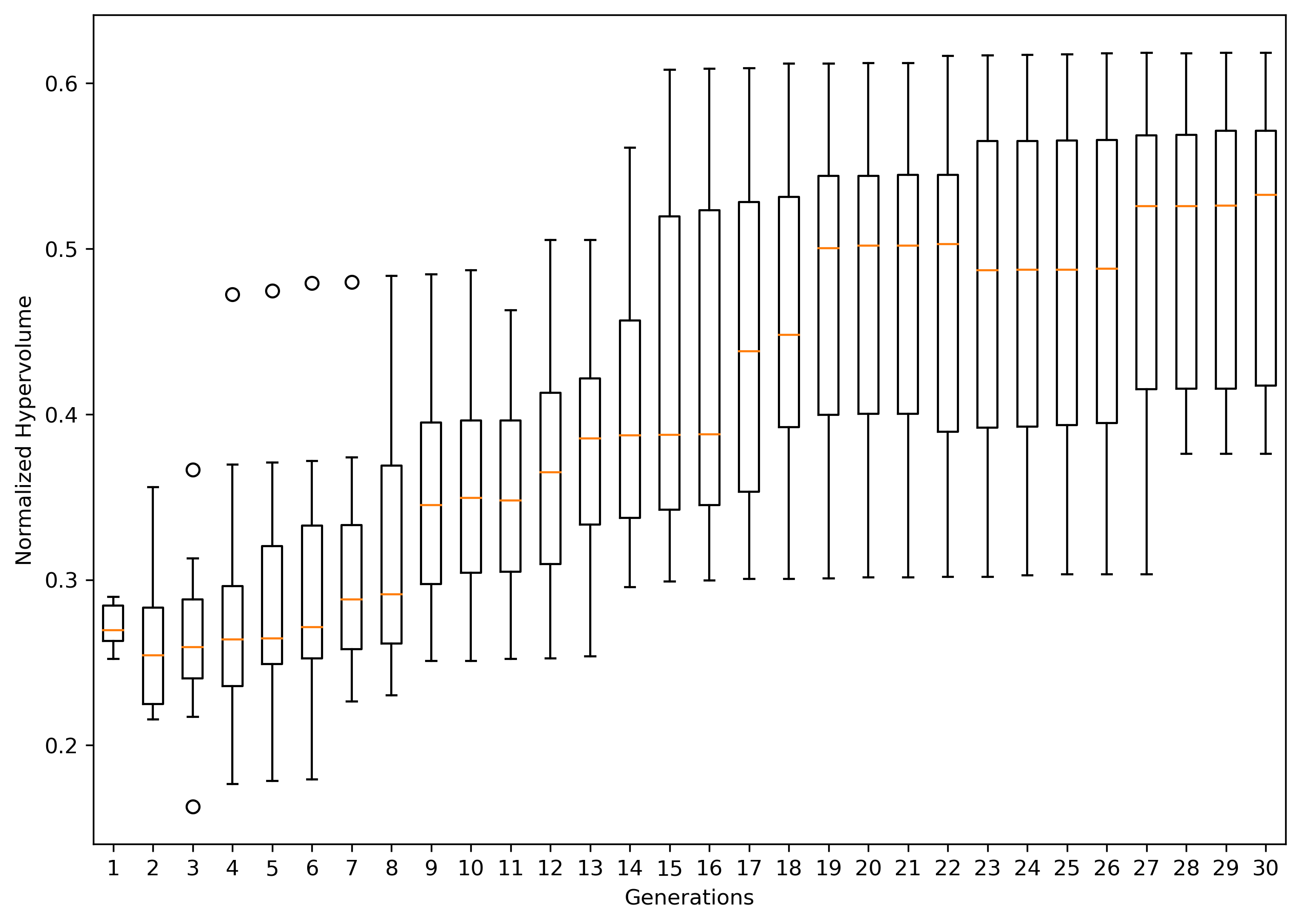}
         \caption{Hypervolume CGPNAS-V1 on CIFAR-100}
         \label{fig:c100hv1}
     \end{subfigure}
      \begin{subfigure}[b]{0.5\textwidth}
         \centering
         \includegraphics[width=\textwidth]{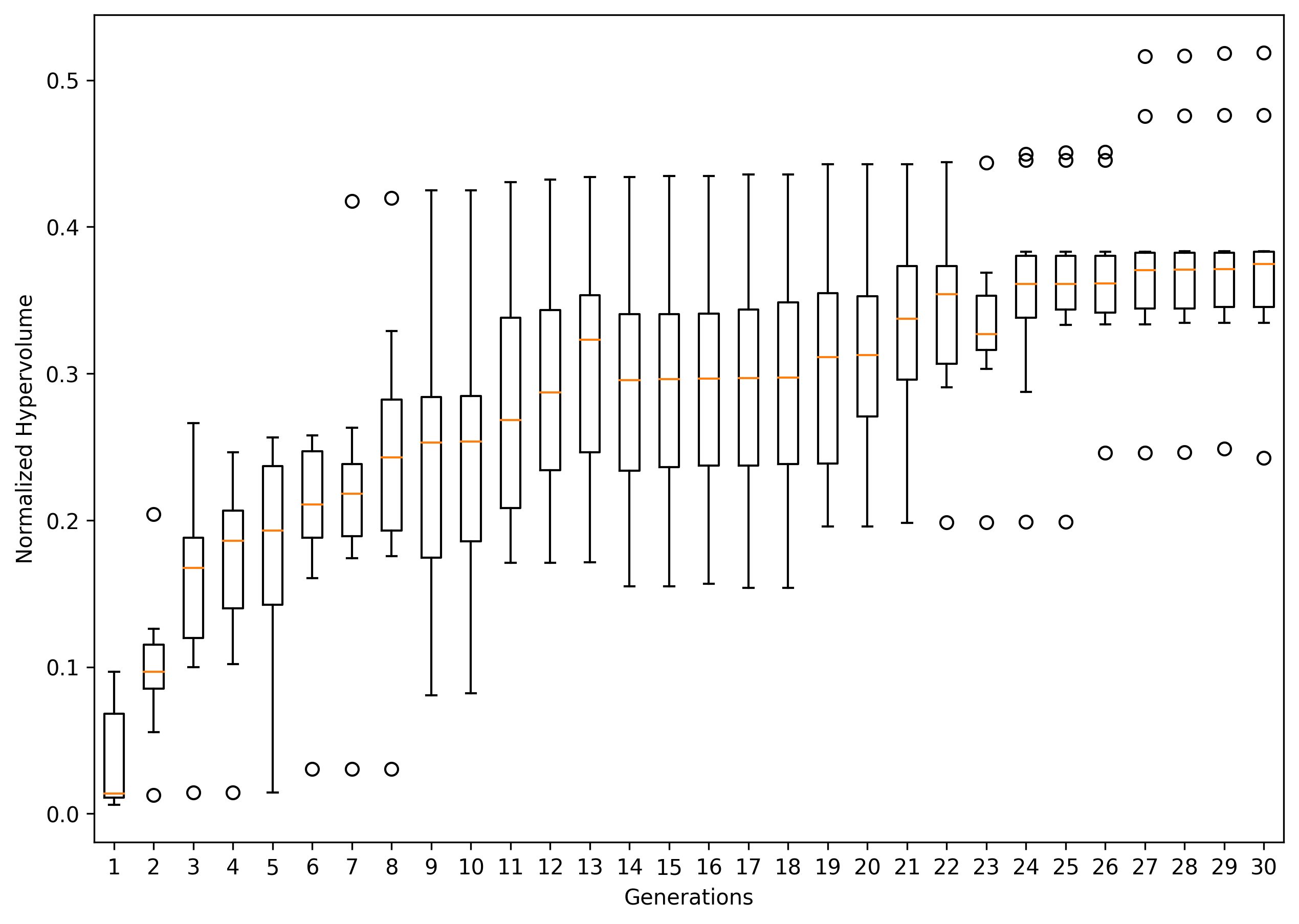}
         \caption{Hypervolume CGPNAS-V2 on CIFAR-100}
         \label{fig:fc100hv2}
     \end{subfigure}

        \caption{Box plot of normalized Hypervolume on CIFAR-10 and CIFAR-100}
        \label{fig:HV}
\end{figure}

Another way to measure the performance of the proposed algorithms is through the Hypervolume metric. It determines the solutions distribution at the Pareto front which means how well they are distributed throughout the objective function space. The Hypervolume metric measures the total area the Pareto front covers with respect to a reference point, we use the nadir point~\citep{Deb2001}. This metric is calculated after every generation and the larger the value, the better. 
In Figure \ref{fig:HV} four box plots are observed each one represents the Hypervolume average value for every generation after $10$ execution samples for the CIFAR-10 and CIFAR-100 datasets.


The final Hypervolume for CGP-NASV1 on CIFAR-10 is  \textbf{0.85(0.651 $\pm$  0.147)}, and for CIFAR-100 it is  \textbf{0.618( 0.505 $\pm$  0.08)}. On the other hand, the results for CGP-NASV2 on CIFAR-10 are  \textbf{0.69(0.551 $\pm$ 0.10)}, and for CIFAR-100, it is CIFAR-100  \textbf{0.51(0.37 $\pm$ 0.07)}. 

From these results and the previous figures, we can observe that in the CIFAR-10 dataset, CGP-NASV1 tends to have more evenly distributed data, as indicated by the length of the whiskers and the size of the boxes, indicating a higher degree of variability in the data between experiments. Conversely, CGP-NASV2 displays less variability in the data, resulting in a more concentrated distribution. On the CIFAR-100 dataset, we can see a similar trend where CGP-NASV1 tends to have more evenly distributed data, while CGP-NASV2 shows a more concentrated distribution and more similar solutions between experiments; however, in the final generations, outliers are generated, which can be caused by sudden and unpredictable changes in the Pareto front resulting from the random nature of the searching algorithm; CGP-NASV2 also influences this behavior because it adds hyperparameters to the search. Summarizing, one can conclude that CGP-NASV1 tends to favor exploration, while CGP-NASV2 favors exploitation at the end of the search processes.

\begin{figure}[h]
    
     \begin{subfigure}[b]{0.5\textwidth}
             \centering
    \includegraphics[scale=0.3]{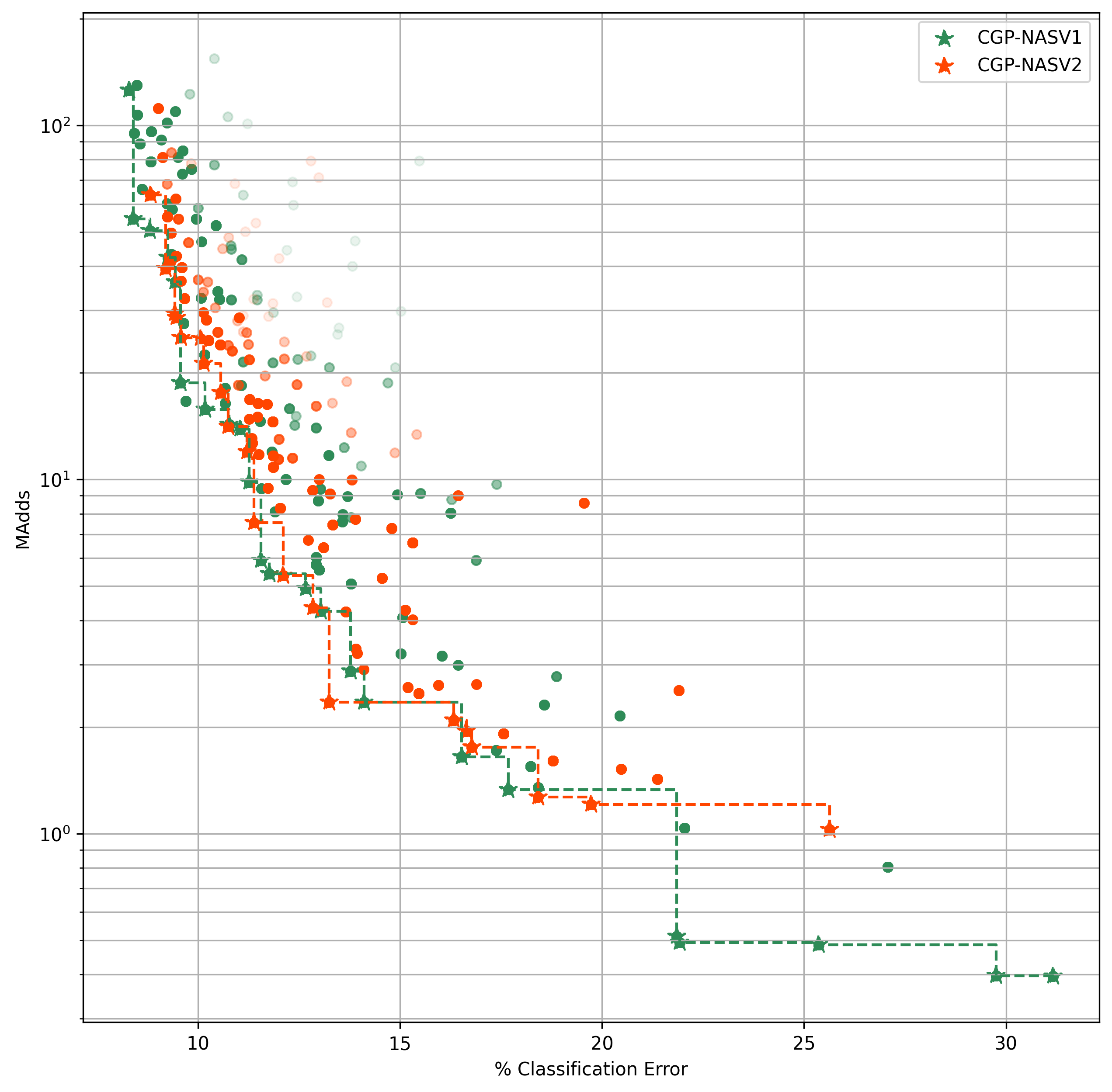}
    \caption{CIFAR-10 Pareto front}
    \label{fig:v1vsv2BC10}
     \end{subfigure}
     \begin{subfigure}[b]{0.5\textwidth}
          \centering
    \includegraphics[scale=0.3]{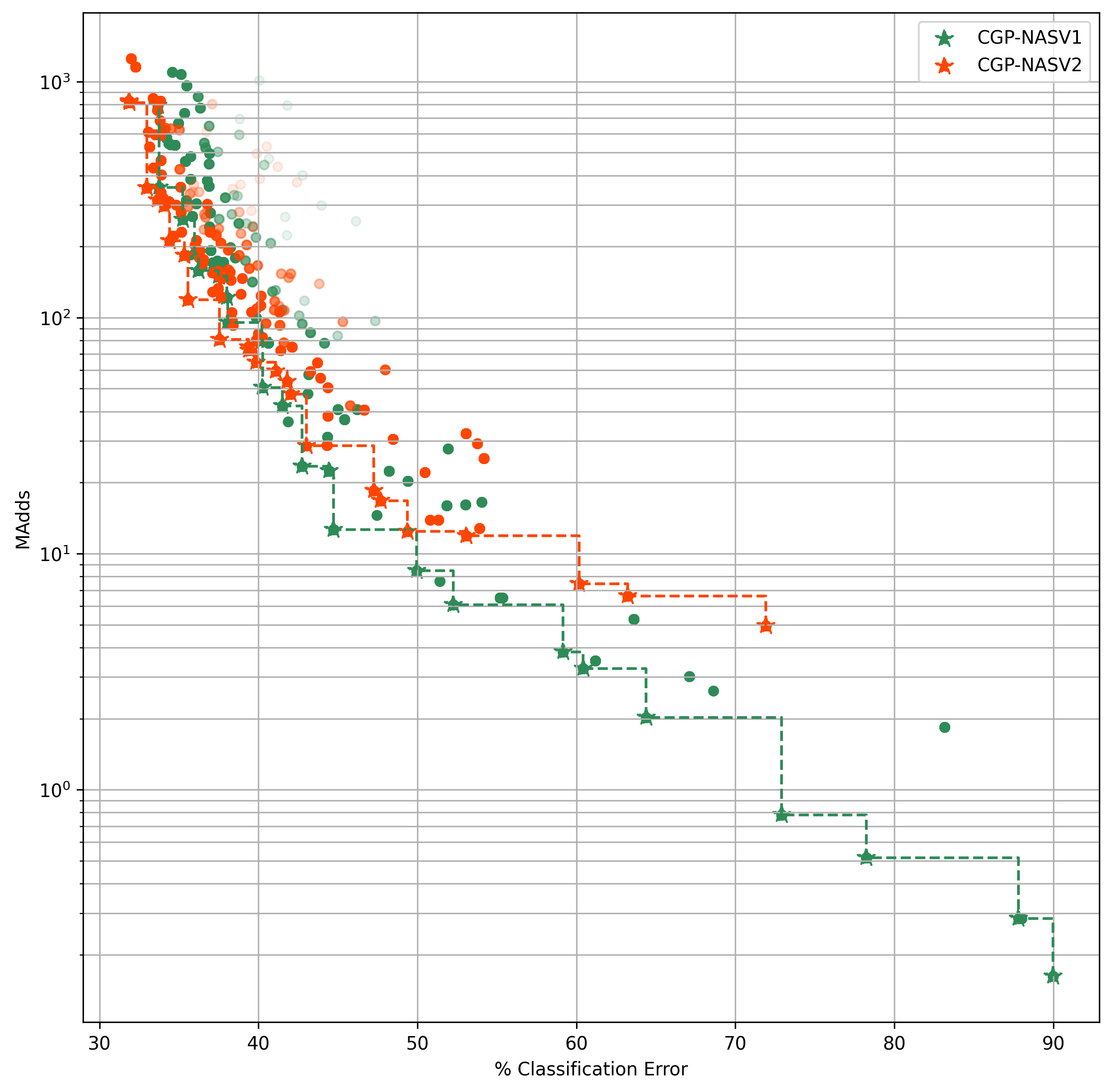}
    \caption{CIFAR-100 Pareto front}
    \label{fig:v1vsv2BC100}
     \end{subfigure}

        \caption{Comparison of the pareto fronts between CGP-NASV1 and CGP-NASV2 , and the population through generations}
        \label{fig:v1v2}
\end{figure}

The search process of CGP-NASV1 and CGP-NASV2 on the CIFAR-10 and CIFAR-100 datasets is illustrated in Figures \ref{fig:v1vsv2BC10} and \ref{fig:v1vsv2BC100}, respectively. The transparency of the circular marks in the figures denotes the relative early stages of the generations. The Pareto front for each proposal is depicted by the corresponding lines. Initially, it can be noted that the solutions are primarily located in the upper left zone, indicating more complex solutions. As the generations progress, solutions are placed in areas further to the right. The region with the most optimal trade-off is located in the lower left corner.

It is also apparent from the figures that the Pareto front for CGP-NASV1  tends to be more disperse in the objective function space. 
CGP-NASV2 demonstrates a tendency to focus on a more restricted area of the objective function space, typically yielding to improve initial solutions when compared to CGP-NASV1

In comparison with other methods, particularly those designed by humans, our proposal demonstrates superior performance in terms of both classification error and number of parameters in both datasets. When compared with single-objective methods, our proposal outperforms them in terms of classification error while showing a significant reduction in the number of parameters. It is important to note, however, that minimizing the number of parameters is not the primary objective of our proposal. When comparing with multi-objective methods in terms of classification error, the EEEA-NET~\citep{Termritthikun2021} method shows better performance, albeit at the cost of a higher number of parameters. In other metrics, our proposal outperforms the other methods presented. Our proposal aims to find solutions with a favorable trade-off between classification error and MAdds, resulting in architectures with reduced parameters and MAdds. In comparison with other proposals that use CGP as a method for architectures representation, such as CGP-CNN~\citep{Suganuma2020}, Torabi~\citep{Torabi2022}, EvoApproxNas~\citep{Pinos123} and LF-MOGP~\citep{Liu2022} , our proposal demonstrates superior performance. 

Works like EvoApproxNAS~\citep{Pinos123} focuses on optimizing values at the hardware level with a multi-objective approach, keeping the CGP representation intact and adding some block functions like bottleneck and residual inverted to the function set while keeping the CGP properties unchanged by only using mutations.
On the other hand, none of these proposals modifies CGP at the representation level, instead attacking specific problems, for example by proposing new operators (Torabi~\citep{Torabi2022}) or mechanisms to improve the search process (LF-MOGP~\citep{Liu2022}), which seems to be an important component since the results shown improve compared to other proposals, even though CGP-NASV2 shows a superior performance with canonical MOEAS.

CGP is known to be a robust method for architecture representation, and the use of real-based representation in our proposal improves the performance, which can be attributed to the relaxation of the search space and the better use of MOEAs, which are designed to operate in a continuous domain.

In summary, CGP-NASV1 and CGP-NASV2 demonstrate the ability to identify solutions with a favorable trade-off between two objectives, specifically by achieving a lower number of MAdds when compared to other methods. It is important to note that depending on the specific application or task at hand, different methods for selecting solutions, such as those presented in this work, may be utilized to determine the most appropriate solution for the given scenario.

\subsection{Use of alternative MOEAs}
To further test the effectiveness of the proposed representation, three additional experiments were conducted. In the first experiment, the differential evolution crossover operator (DE) instead of SBX in NSGA-II was implemented.
In a second and third experiment, MOEA/D and SMS-EMOA were implemented as the searching techniques.
Using MOEA/D and SMS-EMOA as the optimization algorithms would show the adaptability of the proposed CGP-based solution representation to different evolutionary searching strategies.


The experiments were run 10 times on the CIFAR-100 dataset. The DE operator was configured with CR = 1 and F = 0.5, while MOEA/D and SMS-EMOA used the same crossover and mutation mechanisms of NSGA-II and the configurable parameter to define  the number of neighborhoods was set equal $4$. The rest of the parameters were configured the same as those listed in Table \ref{tab:confcgp}.

\begin{table}[h]
\centering
\tiny
\caption{Comparison of CGP-NAS versions on the CIFAR-100 dataset, considering both methods for selecting solutions from the Pareto front. Classification error rate, the parameters and MAdds are expressed in millions ($1 \times 10^{6}$),}
\label{tab:RCGPNASMOEADDE}
\begin{tabular}{|c|ccc|ccc|}
\hline
 &
  \multicolumn{3}{c|}{Knee} &
  \multicolumn{3}{c|}{Best} \\ \hline
 &
  \multicolumn{1}{c|}{Error rate (\%)} &
  \multicolumn{1}{c|}{Parameters} &
  MAdds &
  \multicolumn{1}{c|}{Error rate (\%)} &
  \multicolumn{1}{c|}{Parameters} &
  MAdds \\ \hline
\begin{tabular}[c]{@{}c@{}}CGP-NASV2\\ \textbf{NSGA-II-DE}\end{tabular} &
  \multicolumn{1}{c|}{\begin{tabular}[c]{@{}c@{}}24.75\\ (27.15 ± 1.85)\end{tabular}} &
  \multicolumn{1}{c|}{\begin{tabular}[c]{@{}c@{}}1.09\\ (0.78 + 0.21)\end{tabular}} &
  \begin{tabular}[c]{@{}c@{}}86.49\\ (71.06+ 12.03)\end{tabular} &
  \multicolumn{1}{c|}{\begin{tabular}[c]{@{}c@{}}21.02\\ (22.66± 0.99)\end{tabular}} &
  \multicolumn{1}{c|}{\begin{tabular}[c]{@{}c@{}}5.99\\ (5.89 + 2.75)\end{tabular}} &
  \begin{tabular}[c]{@{}c@{}}960.01\\ (1164.75 + 559.59)\end{tabular} \\ \hline
  
\begin{tabular}[c]{@{}c@{}}CGP-NASV2\\\textbf{MOEA/D}\end{tabular} &
  \multicolumn{1}{c|}{\begin{tabular}[c]{@{}c@{}}29.47\\ (32.46 ± 1.42)\end{tabular}} &
  \multicolumn{1}{c|}{\begin{tabular}[c]{@{}c@{}}0.32\\ (0.34 ± 0.07)\end{tabular}} &
  \begin{tabular}[c]{@{}c@{}}36.71\\ (30.75± 7.70)\end{tabular} &
  \multicolumn{1}{c|}{\begin{tabular}[c]{@{}c@{}}21.12\\ (23.88 ± 1.7)\end{tabular}} &
  \multicolumn{1}{c|}{\begin{tabular}[c]{@{}c@{}}5.30\\ (3.98 ± 1.11)\end{tabular}} &
  \begin{tabular}[c]{@{}c@{}}1021.62\\ (601.18± 186.12)\end{tabular} \\ \hline
  
\begin{tabular}[c]{@{}c@{}}CGP-NASV2\\ \textbf{NSGA-II-SBX}\end{tabular} & 
  \multicolumn{1}{c|}{\textbf{\begin{tabular}[c]{@{}c@{}}23.57\\ (28.43± 2.20)\end{tabular}}} &
  \multicolumn{1}{c|}{\begin{tabular}[c]{@{}c@{}}0.49\\ (0.53 ±0.13)\end{tabular}} &
  \begin{tabular}[c]{@{}c@{}}66.66\\ (55.66 ±19.14)\end{tabular} &
  \multicolumn{1}{c|}{\textbf{\begin{tabular}[c]{@{}c@{}}20.63\\ (22.49 ± 1.04)\end{tabular}}} &
  \multicolumn{1}{c|}{\begin{tabular}[c]{@{}c@{}}5.99\\ (6.50 ± 1.7)\end{tabular}} &
  \begin{tabular}[c]{@{}c@{}}827\\ (850.74± 476.08)\end{tabular} \\ \hline
  
\begin{tabular}[c]{@{}c@{}}CGP-NASV2\\ \textbf{SMS-EMOA}\end{tabular} & 
  \multicolumn{1}{c|}{\begin{tabular}[c]{@{}c@{}}25.38\\ (26.84 ± 1.68)\end{tabular}} &
  \multicolumn{1}{c|}{\begin{tabular}[c]{@{}c@{}}0.95\\ (0.54 ± 0.19)\end{tabular}} &
  \begin{tabular}[c]{@{}c@{}}98.74\\ (61.09± 17.84)\end{tabular} &
  \multicolumn{1}{c|}{\begin{tabular}[c]{@{}c@{}}21.55\\ (22.82 ± 1.06)\end{tabular}} &
  \multicolumn{1}{c|}{\begin{tabular}[c]{@{}c@{}}12.19\\ (7.10 ± 3.21)\end{tabular}} &
  \begin{tabular}[c]{@{}c@{}}1320.17\\ (846.15± 261.03)\end{tabular} \\ \hline
\end{tabular}
\end{table}

Table \ref{tab:RCGPNASMOEADDE} shows the results for CGP-NASV2 with different evolutionary strategies after $10$ executions on the CIFAR-100 dataset. Our experiments revealed several interesting behaviors. In the canonical version of CGP-NASV2 with NSGA-II and SBX , we consistently obtained the best classification errors. When using the DE operator, we saw similar results to CGP-NASV2, with a slightly lower average and standard deviation, indicating more consistent behavior. Finally, MOEA/D performed similarly to DE while producing solutions with significantly lower complexity, potentially due to its decomposition approach. In the case of SMS-EMOA, we see a very similar performance to NSGA-II-SBX. However,  we can notice that on average, in SMS-EMOA, the solutions have a lower or equal error than NSGA-II-SBX. This may be due to the indicator-based approach of SMS-EMOA, but the best solution located is worse than our baseline.

To further illustrate these findings, we present the Pareto fronts for each method below, including the average Pareto front for the 10 runs per method. These graphs demonstrate the evolution of the trade-off between classification error and complexity over the course of several generations



\begin{figure}
    
     \begin{subfigure}[b]{0.5\textwidth}
         \centering
         \includegraphics[scale=0.31]{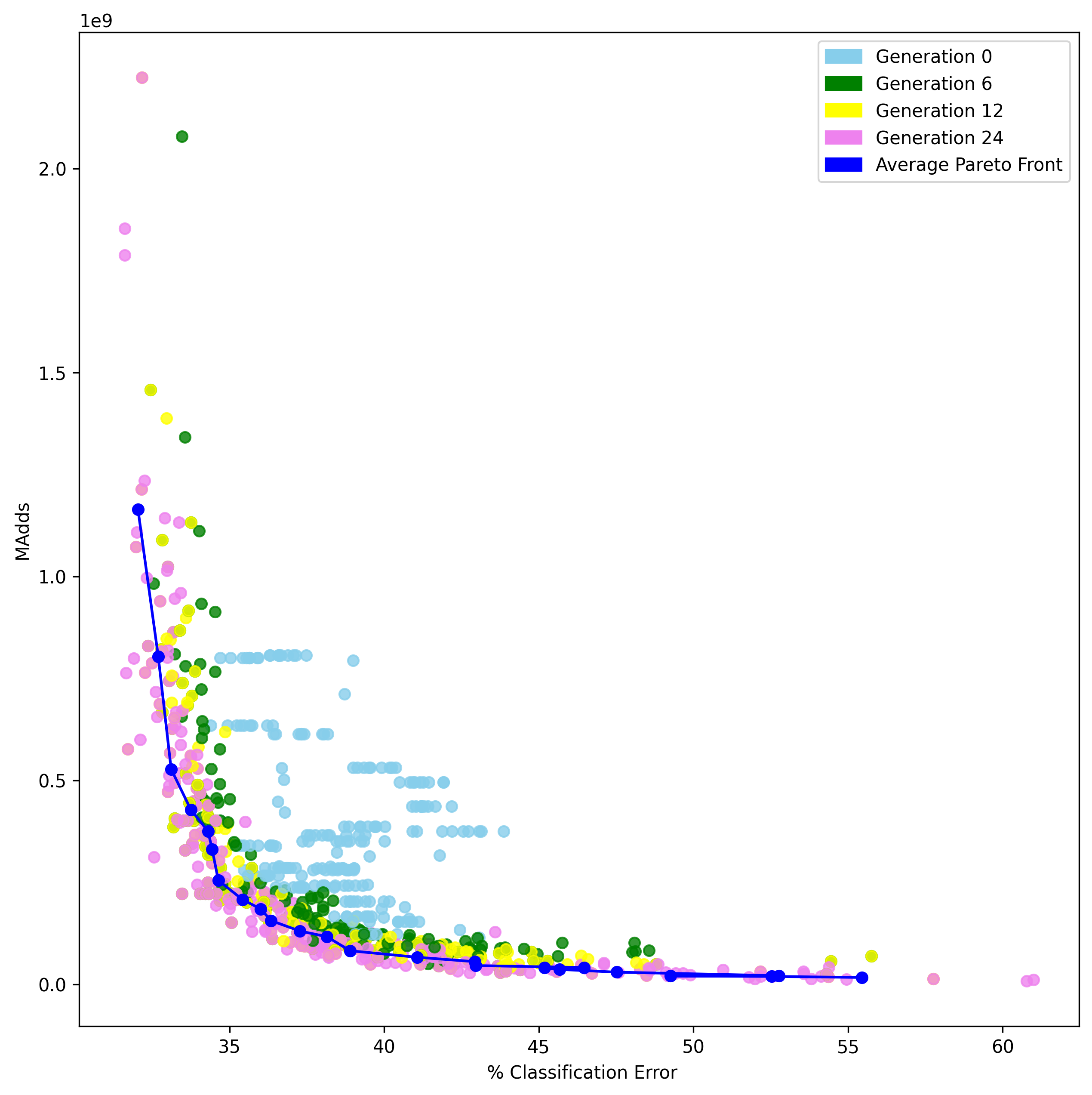}
         \caption{CGP-NASV2 with NSGA-II-DE}
         \label{fig:rcgpDE}
     \end{subfigure}
     \begin{subfigure}[b]{0.5\textwidth}
         \centering
         \includegraphics[scale=0.31]{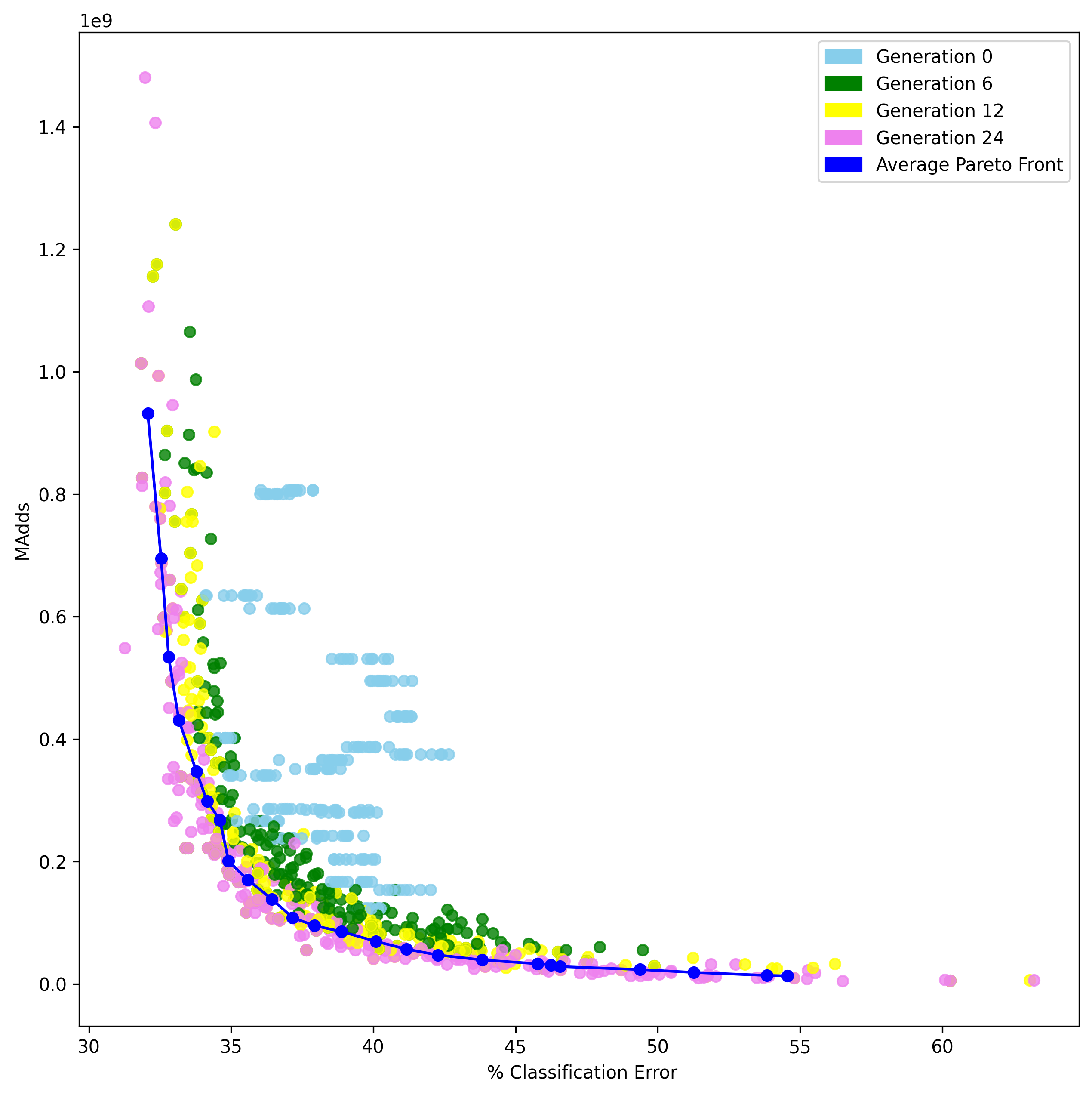}
         \caption{CGP-NASV2 with NSGA-II-SBX}
         \label{fig:rcgpnsga}
     \end{subfigure}
     
        \caption{Evolution progress of CGP-NASV2-SBX and CGP-NASV2-DE in the CIFAR100 dataset, each color refers to a specific generation.}
        \label{fig:SBXDE}
\end{figure}

\begin{figure}
    
     \begin{subfigure}[b]{0.5\textwidth}
         \centering
         \includegraphics[scale=0.31]{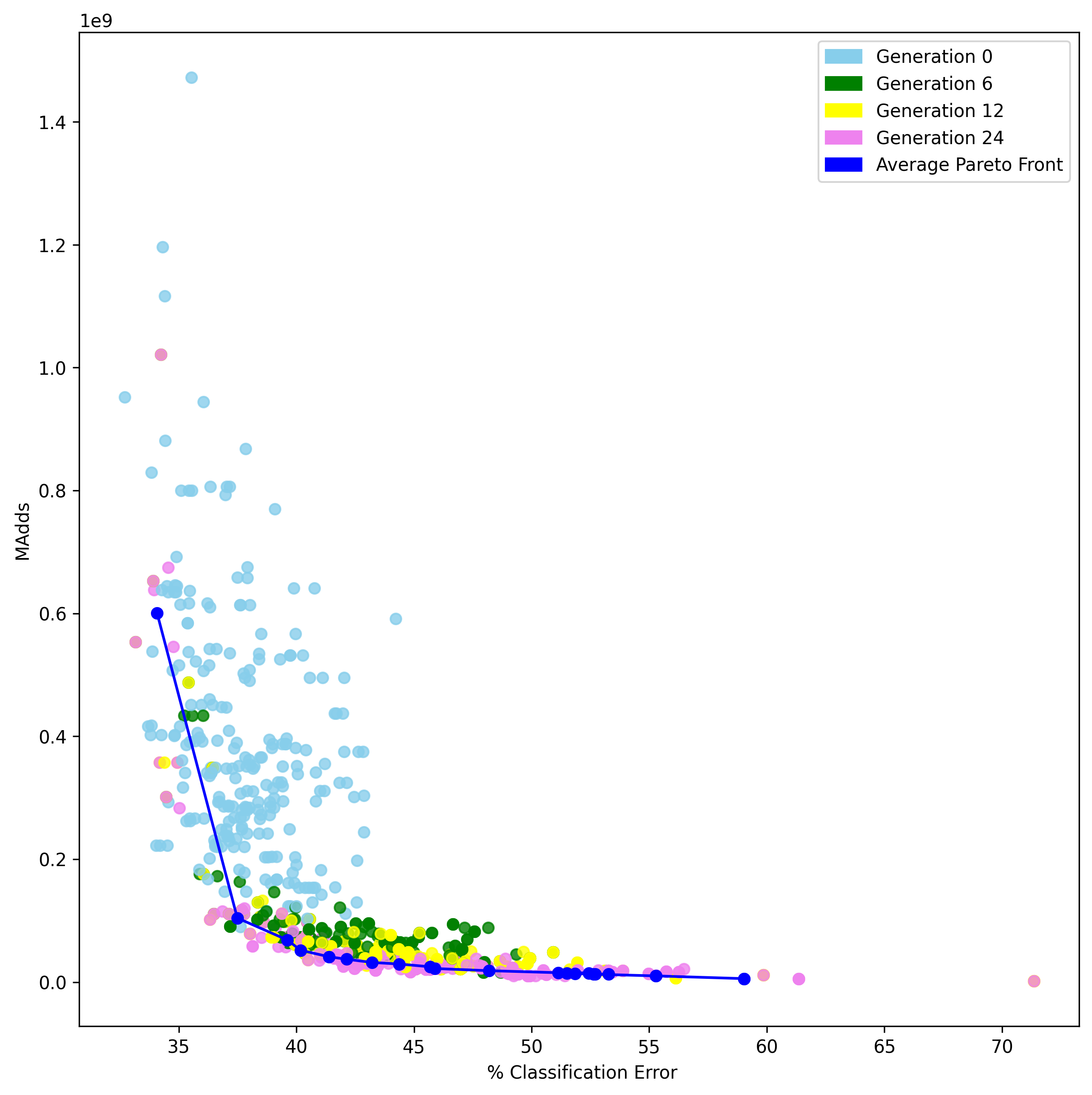}
         \caption{CGP-NASV2 with MOEA/D}
         \label{fig:rcgpMD}
     \end{subfigure}
     \begin{subfigure}[b]{0.5\textwidth}
         \centering
         \includegraphics[scale=0.31]{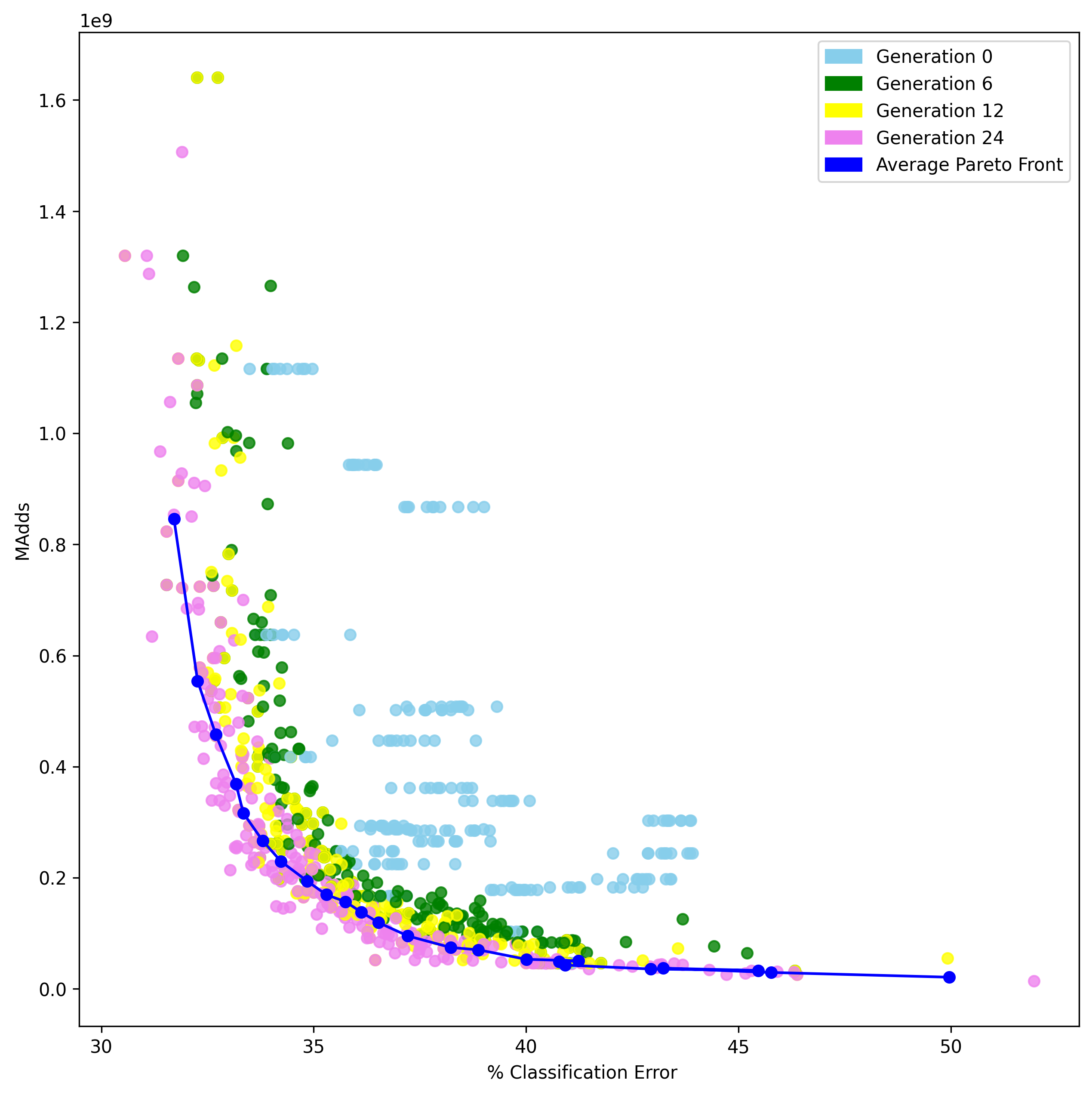}
         \caption{CGP-NASV2 with SMS-EMOA}
         \label{fig:rcgpSMS}
     \end{subfigure}
     
        \caption{Evolution progress of CGP-NASV2 with SMS-EMOA and MOEA/D core algorithms in the CIFAR100 dataset, each color refers to a specific generation.}
        \label{fig:MDSMS}
\end{figure}



One striking feature of the Pareto fronts is the strong similarity between Figures \ref{fig:rcgpnsga} and \ref{fig:rcgpDE}, which use the same MOEA. As expected, we see that the DE operator leads to more complex solutions, while MOEA/D exhibits a decrease in MAdds in the early generations, possibly due to the fast convergence of the algorithm, while still maintaining the trade-off in later generations. 
On the other hand, SMS-EMOA also shares quite a large similarity with NSGA-II-SBX, although we see better distributed solutions across generations because, in the background, SMS-EMOA keeps the solutions that improve the hypervolume.

Overall, the results show the impact of the different mechanisms offered by each MOEA and the impact they have on the trade-off between classification error and complexity in the multi-objective neural architecture search.

\subsection{Discussion}

\begin{figure}
    
     \begin{subfigure}[b]{0.5\textwidth}
         \centering
         \includegraphics[scale=0.08]{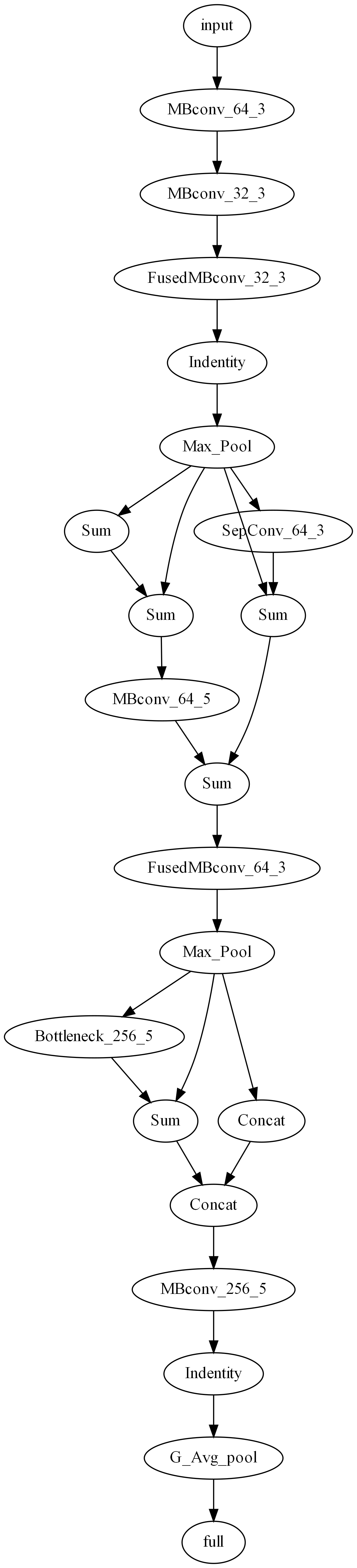}
         \caption{For CIFAR-10 dataset }
         \label{fig:c10AQv1}
     \end{subfigure}
     \begin{subfigure}[b]{0.5\textwidth}
         \centering
         \includegraphics[scale=0.08]{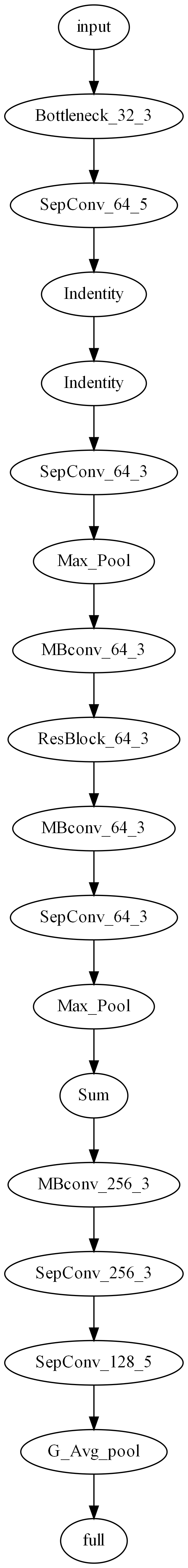}
         \caption{For CIFAR-100 dataset}
         \label{fig:c10AQv2}
     \end{subfigure}

        \caption{CGP-NASV2 evolved CNN architectures selected by the knee and boundary method.}
        \label{fig:ARQCNN}
\end{figure}

Figure~\ref{fig:ARQCNN} show two CGP-NASV2 evolved architectures selected through the knee and boundary method. Figure~\ref{fig:c10AQv1} correspond to the CIFAR-10 dataset; it is observed that the architecture has more branches together with parameter-free blocks such as summation and concatenation blocks. Figure~\ref{fig:c10AQv2} for the CIFAR-100 dataset, shows an architecture with a more linear structure but with more complex blocks such as MBConv and SepConv. After these observations it is possible to assume that CGP-NASV2 on the more difficult CIFAR-100 data set, evolved by selecting the necessary blocks (even if they are more complex) without significantly increasing the number of parameters or MAdds and therefore maintaining a low classification error. 


Comparing the best proposed performing NAS algorithm, CGP-NASV2 versus the previous proposal CGP-NAS~\citep{Garcia-Garcia2022} using the same selection method, there was an improvement from 4.86\% to 3.70\% in the classification error. However, there is also an increase in the number of parameters and MAdds, that can be reduced by selecting an evolved architecture with the best trade-off between objectives. Thus a solution with similar performance but with much lower complexity is also available. Similar analysis applies to those evolved architectures for the CIFAR-100 dataset.


\label{c}
\section{Conclusion}
This study presents a multi-objective evolutionary approach for neural architecture search (NAS) applied to image classification tasks using the CIFAR-10 and CIFAR-100 datasets. The proposed methods, CGP-NASV1 and CGP-NASV2, were able to find CNN architectures that balance classification error and complexity, as measured by the number of MAdds. A key advantage of CGP is its flexibility, allowing for the exploration of different evolutionary searching strategies and operators.

The results demonstrate that the CGP-NASV2 method is effective in finding CNN architectures with fewer parameters and MAdds compared to other state-of-the-art methods. Additionally, we conducted an exploratory study to investigate the effects of using different multiobjective evolutionary searching strategies and operations. By replacing the crossover operator with the DE operator and using other MOEAs like MOEA/D and SMS-EMOA, we note that our CGP-based representation brings multiple benefits, from allowing variable-length architectures to the adaptation of multiple operators and search schemes due to the fact that our representation works in a real domain. In summary, the study demonstrates the effectiveness and flexibility of the CGP-based representation for NAS.

\section*{Acknowledgements}
The authors thankfully acknowledge computer resources, technical advice and support provided by Laboratorio Nacional de Supercómputo del Sureste de México (LNS), a member of the CONACYT national laboratories, with project No. 202103083C.  This work was supported by CONACyT under grant CB-S-26314. 



 \bibliographystyle{elsarticle-num-names} 
 \bibliography{MAIN.bib}





\end{document}